\documentclass[11pt]{article}
\usepackage[table,xcdraw]{xcolor}
\usepackage[a4paper,margin=1in]{geometry}
\usepackage{graphicx}
\usepackage{amsmath,amssymb}
\usepackage{hyperref}
\usepackage{tcolorbox}
\usepackage{tabularray}
\usepackage{subcaption}
\usepackage{multirow}
\usepackage[normalem]{ulem}
\useunder{\uline}{\ul}{}

\newcommand{\redtext}[1]{\textcolor{red}{#1}}
\renewcommand{\redtext}[1]{#1} 

\newlength{\myrowsep}
\newlength{\myrowsepsmall}
\title{LLM4TS: Aligning Pre-Trained LLMs as Data-Efficient Time-Series Forecasters}
\author{
  Ching Chang\textsuperscript{1}, Wei-Yao Wang\textsuperscript{1}, Wen-Chih Peng\textsuperscript{1}, Tien-Fu Chen\textsuperscript{1} \\
  \textsuperscript{1}National Yang Ming Chiao Tung University, Hsinchu, Taiwan \\
  \texttt{blacksnail789521.cs10@nycu.edu.tw, sf1638.cs05@nctu.edu.tw,} \\ 
  \texttt{wcpeng@cs.nycu.edu.tw, tfchen@cs.nycu.edu.tw}
}

\date{}  

\begin{document}

\begin{tcolorbox}[colback=gray!10, colframe=black, title=Preprint Notice]
This paper has been accepted for publication in ACM Transactions on Intelligent Systems and Technology (TIST) 2025. 
The final version will be available at \url{https://doi.org/10.1145/3719207}.
\end{tcolorbox}

\maketitle

\begin{abstract}
Multivariate time-series forecasting is vital in various domains, e.g., economic planning and weather prediction.
Deep train-from-scratch models have exhibited effective performance yet require large amounts of data, which limits real-world applicability.
Recently, researchers have leveraged the representation learning transferability of pre-trained Large Language Models (LLMs) to handle limited non-linguistic datasets effectively.
However, incorporating LLMs with time-series data presents challenges of limited adaptation due to different compositions between time-series and linguistic data, and the inability to process multi-scale temporal information.
To tackle these challenges, we propose LLM4TS, a framework for time-series forecasting with pre-trained LLMs.
LLM4TS consists of a two-stage fine-tuning strategy: the \textit{time-series alignment} stage to align LLMs with the nuances of time-series data, and the \textit{forecasting fine-tuning} stage for downstream time-series forecasting tasks.
Furthermore, our framework features a novel two-level aggregation method that integrates multi-scale temporal data within pre-trained LLMs, enhancing their ability to interpret time-specific information.
In experiments across 7 time-series forecasting datasets, LLM4TS is superior to existing state-of-the-art methods compared with trained-from-scratch models in full-shot scenarios, and also achieves the highest rank in few-shot scenarios.
In addition, evaluations compared with different unsupervised representation learning approaches highlight LLM4TS's effectiveness with representation learning in forecasting tasks.
Ablation studies further validate each component's contribution to LLM4TS and underscore the essential role of utilizing LLM's pre-trained weights for optimal performance.
The code is available at \url{https://github.com/blacksnail789521/LLM4TS}.
\end{abstract}

\section{Introduction}
Forecasting is a vital task in multivariate time-series analysis, not only for its ability to operate without manual labeling but also for its importance in practical applications such as economic planning \cite{exchange_rate_1, exchange_rate_2, exchange_rate_3} and weather prediction \cite{informer, weather_prediction_1, weather_prediction_2}.
Recently, numerous deep train-from-scratch models have been developed for time-series forecasting \cite{dlinear, patchtst, autoformer, fedformer, informer, forecasting_extra}, although some lean towards unsupervised representation learning \cite{btsf, ts2vec, tnc, ts_tcc, simts, timedrl, unsupervsed_ts_survey} and transfer learning \cite{tf_c, gpt4ts, transfer_learning_1, transfer_learning_2, transfer_learning_3}.
Generally, these approaches aim to employ adept representation learners: first extracting rich representations from the time-series data and then using these representations for forecasting.

Achieving an adept representation learner requires sufficient training data \cite{chinchilla, need_big_data_1, need_big_data_2}, yet in real-world scenarios, there is often a lack of large-scale time-series datasets.
For instance, in industrial manufacturing, the sensor data for different products cannot be combined for further analysis, leading to limited data for each product type \cite{industry_1, industry_2, industry_3}.
Recent research has pivoted towards pre-trained LLMs in Natural Language Processing (NLP) \cite{gpt2, gpt3, llama}, exploiting their robust representation learning and few-shot learning capabilities.
Moreover, these LLMs can adapt to non-linguistic datasets (e.g., images \cite{fpt, llava},  audio \cite{tango, audio_llm_extra}, tabular data \cite{tabllm, tabular_llm_extra}, and time-series data \cite{gpt4ts, timellm}) by fine-tuning with only a few parameters and limited data.
While LLMs are renowned for their exceptional transfer learning capabilities across various fields, the domain-specific nuances of time-series data introduce two challenges in leveraging these models for time-series forecasting.

The first challenge of employing LLMs for time-series forecasting is their limited adaptation to the unique characteristics of time-series data due to LLMs' initial pre-training focus on the linguistic corpus.
While LLMs have been both practically and theoretically proven \cite{gpt4ts} to be effective in transfer learning across various modalities thanks to their data-independent self-attention mechanism, their primary focus on general text during pre-training causes a shortfall in recognizing key time-series patterns and nuances crucial for accurate forecasting.
This limitation is evident in areas such as meteorology and electricity forecasting \cite{informer}, where failing to account for weather patterns and energy consumption trends leads to inaccurate predictions.

The second challenge lies in the limited capacity to process multi-scale temporal information.
While LLMs are adept at understanding the sequence and context of words, they struggle to understand temporal information due to the lack of utilizing multi-scale time-related data such as time units (e.g., seconds, minutes, hours, etc.) and specific dates (e.g., holidays, significant events).
This temporal information is vital in time-series analysis for identifying and predicting patterns \cite{autoformer, timesnet}; for instance, in energy management, it is used to address consumption spikes during daytime and in summer/winter, in contrast to the lower demand during the night and in milder seasons \cite{informer}.
This underscores the importance of models adept at interpreting multi-scale temporal patterns (hourly to seasonal) for precise energy demand forecasting.
However, most LLMs (e.g., \cite{gpt2,llama}) built on top of the Transformer architecture do not naturally incorporate multi-scale temporal information, leading to models that fail to capture crucial variations across different time scales.

\begin{figure}[t]
    \centering
    \includegraphics[width=1\linewidth]{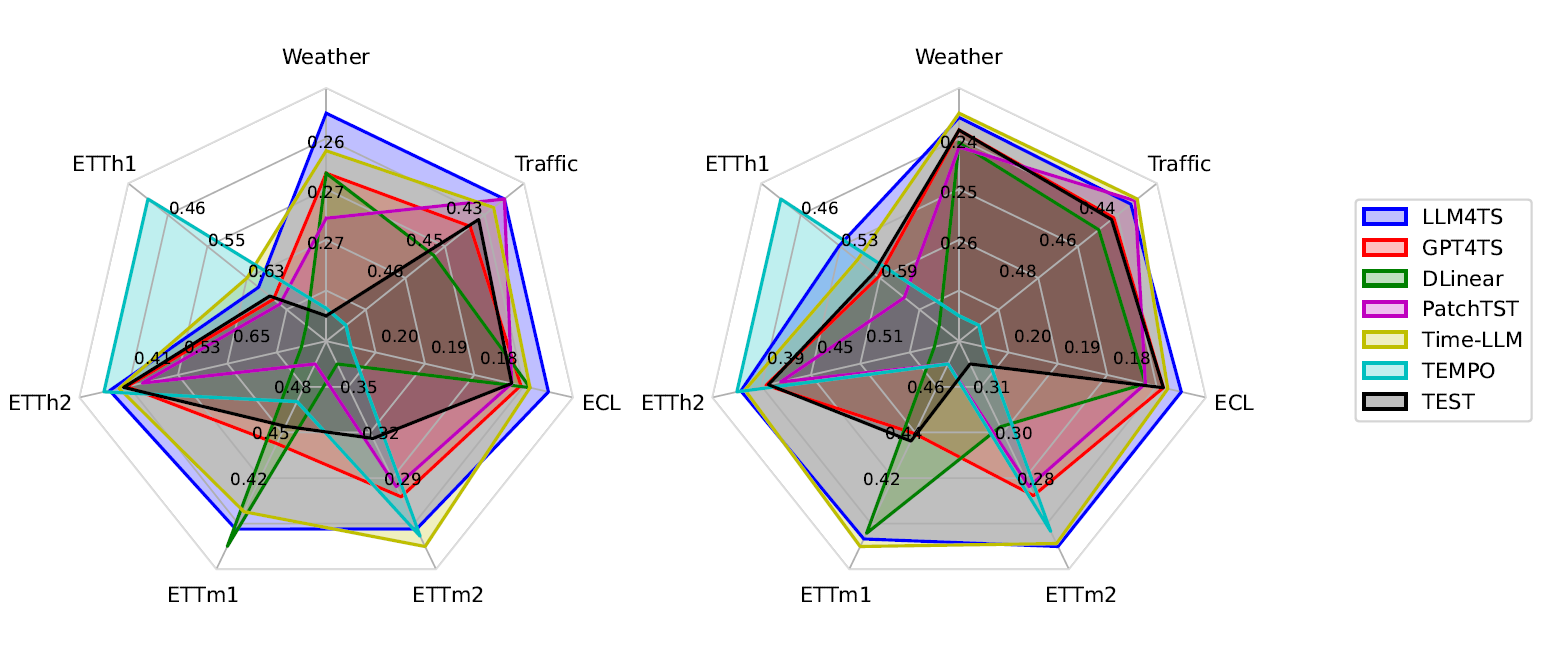}

    \begin{subfigure}{.3\linewidth}
        \centering
        \caption{5\% training data}
    \end{subfigure}
    \hspace{1.4cm}
    \begin{subfigure}{.3\linewidth}
        \centering
        \caption{10\% training data}
    \end{subfigure}
    \hspace{2.5cm}
    \caption{\redtext{\textbf{Model performance comparison on few-shot forecasting.}}}
    \label{fig:pie_chart}
\end{figure}

To address the above issues, we propose LLM4TS, a framework for time-series forecasting with pre-trained LLMs.
Regarding the first challenge, our framework introduces a two-stage fine-tuning approach: the \textit{time-series alignment} stage and the \textit{forecasting fine-tuning} stage.
The first stage focuses on aligning the LLMs with the characteristics of time-series data by utilizing the autoregressive objective, enabling the fine-tuned LLMs to adapt to time-series representations.
The second stage is incorporated to learn corresponding time-series forecasting tasks.
In this manner, our model supports effective performance in full- and few-shot scenarios.
Notably, throughout both stages, most parameters in the pre-trained LLMs are frozen, thus preserving the model's inherent representation learning capability.
To overcome the limitation of LLMs in integrating multi-scale temporal information, we introduce a novel two-level aggregation strategy.
This approach embeds multi-scale temporal information into the patched time-series data, ensuring that each patch not only represents the series values but also encapsulates the critical time-specific context.
Consequently, LLM4TS emerges as a data-efficient time-series forecaster, demonstrating robust few-shot performance across various datasets (Fig. \ref{fig:pie_chart}).

In summary, the paper's main contributions are as follows:
\begin{itemize}
    \item \textbf{Aligning LLMs Toward Time-Series Data:} To the best of our knowledge, LLM4TS is the first method that aligns pre-trained Large Language Models with time-series characteristics, effectively utilizing existing representation learning and few-shot learning capabilities.
    \item \textbf{Multi-Scale Temporal Information in LLMs:} To adapt to time-specific information, a two-level aggregation method is proposed to integrate multi-scale temporal data within pre-trained LLMs.
    \item \textbf{Robust Performance in Forecasting:} LLM4TS excels in 7 real-world time-series forecasting benchmarks, outperforming state-of-the-art methods, including those trained from scratch. It also demonstrates strong few-shot capabilities, particularly with only \(5\%\) of data, where it surpasses the best baseline that uses \(10\%\) of data. This efficiency makes LLM4TS highly relevant for practical, real-world forecasting applications.
\end{itemize}


\section{Related Work}
\subsection{Transfer Learning Across Various Modalities with LLMs}
\label{sec:transfer_learning_with_llms}
LLMs have demonstrated their effectiveness in transfer learning across a variety of modalities, such as images \cite{fpt, llava}, audio \cite{tango, audio_llm_extra}, tabular data \cite{tabllm, tabular_llm_extra}, and time-series data \cite{gpt4ts, timellm}.
A key motivation for employing LLMs in various modalities is their ability to achieve notable performance with limited data \cite{gpt4ts}.
To preserve their data-independent representation learning capability, most parameters in these LLMs are kept fixed.
Empirical evidence \cite{fpt,gpt4ts} indicates that LLMs keeping most parameters unchanged often outperform those trained from scratch, underscoring the value of maintaining these models' pre-existing representation learning strengths (more experiments can be found in Section \ref{sec:ablation_part_c}).
Theoretically, it is shown that the self-attention modules in these pre-trained transformers develop the capacity for data-independent operations (akin to principal component analysis \cite{gpt4ts}), enabling them to function effectively as \textit{universal compute engines} \cite{fpt} or \textit{general computation calculators} \cite{general_computation_calculator}.
In the time-series domain, GPT4TS \cite{gpt4ts} utilizes the pre-trained GPT-2 and demonstrates strong performance in time-series forecasting under few-shot conditions without modifying most parameters.
\redtext{
Time-LLM \cite{time_llm} reprograms LLMs for time series forecasting by converting time series into text prototypes and using prompts to guide prediction. It outperforms specialized models, particularly in few-shot and zero-shot settings.
TEMPO \cite{tempo} adapts GPT-like models for time series forecasting by decomposing trends and using prompts for better distribution adaptation, excelling in zero-shot scenarios.
TEST \cite{test} aligns time series with LLM embeddings through tokenization and contrastive learning, enabling effective time series forecasting using pre-trained LLMs without fine-tuning.
}
With our LLM4TS, we address the challenges of limited adaptation to time-series characteristics and the difficulty in processing multi-scale temporal information, thereby enhancing performance in time-series forecasting.

\subsection{Long-term Time-Series Forecasting}
Numerous efforts have been dedicated to employing Transformer models for long-term time-series forecasting \cite{informer,autoformer,fedformer,patchtst,crossformer,ts_fastformer}.
While Transformer-based models have gained traction, DLinear \cite{dlinear} reveals that a single-layer linear model can surpass many of these sophisticated Transformer-based approaches.
These deep train-from-scratch models exhibit outstanding performance when trained on sufficient datasets, but their efficacy decreases in limited-data scenarios.
In contrast, LLM4TS sets new benchmarks alongside these state-of-the-art approaches in both full- and few-shot scenarios.

\subsection{Time-Series Representation Learning}
In the time-series domain, self-supervised learning emerges as a prominent approach to representation learning.
While Transformers are widely recognized as prime candidates for end-to-end time-series analysis \cite{anomaly_transformer,gtn,patchtst,time_series_transformer_survey_1,time_series_transformer_survey_2}, CNN-based \cite{ts2vec} or RNN-based \cite{tnc} backbones consistently stand out as the preferred architecture in time-series self-supervised learning.
However, the inherent capability of Transformers to model long-range dependencies and capture patterns aligns perfectly with time-series data, which involve complex sequential relationships.
Since the \textit{time-series alignment} stage in LLM4TS can be seen as a self-supervised learning approach, we evaluate LLM4TS's representation learning capability and demonstrate the full potential of Transformers in unsupervised representation learning, surpassing the performance of conventional CNN and RNN-based models.

\begin{figure}[t]
    \centering
    \includegraphics[width=0.545\linewidth]{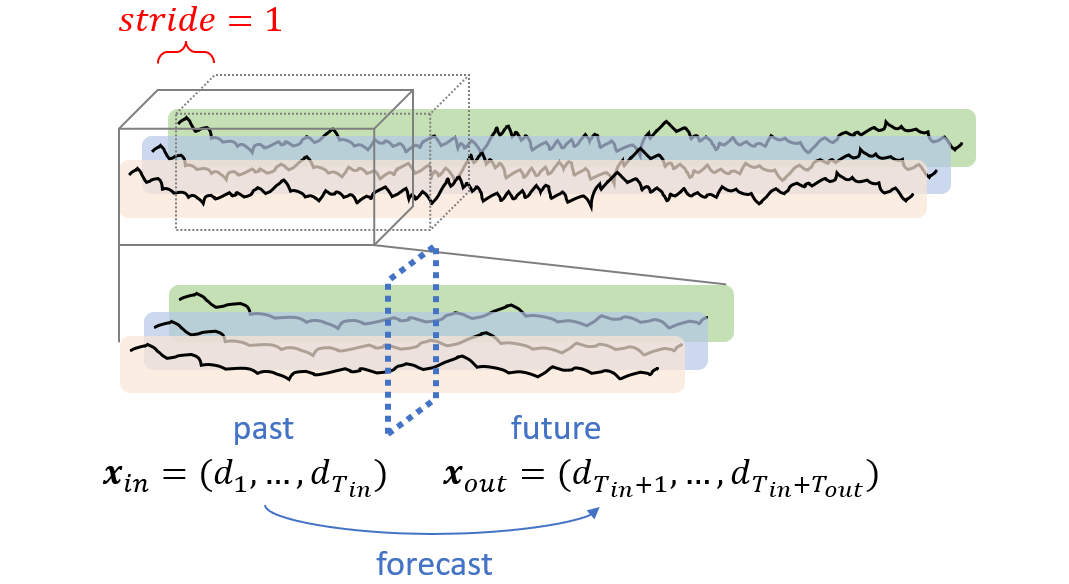}
    \caption{\redtext{\textbf{Problem formulation for multivariate time-series forecasting}.}}
    \label{fig:problem_formulation}
\end{figure}

\section{Problem Formulation}
\redtext{
Given a complete and evenly-sampled multivariate time-series, we use a sliding data window to extract sequential samples, as illustrated in Fig. \ref{fig:problem_formulation}.
}
This window moves with a stride of \(1\) and has a total length of \(T_{in}+T_{out}\) — comprising past data \( \textbf{x}_{in}= (d_1, \dots, d_{T_{in}}) \) with a look-back window length \(T_{in}\) and future data \( \textbf{x}_{out}= (d_{T_{in}+1}, \dots, d_{T_{in}+T_{out}}) \) with a prediction length \(T_{out}\).
For each time step \( t \), \( d_t \) represents a \( C \)-dimensional vector, where \(C\) denotes the number of features.
Our objective is to use the past data \(\textbf{x}_{in} \in \mathbb{R}^{T_{in} \times C}\) to predict the future data \(\textbf{x}_{out} \in \mathbb{R}^{T_{out} \times C}\).

\begin{figure*}[t]
    \centering
    \includegraphics[width=\textwidth]{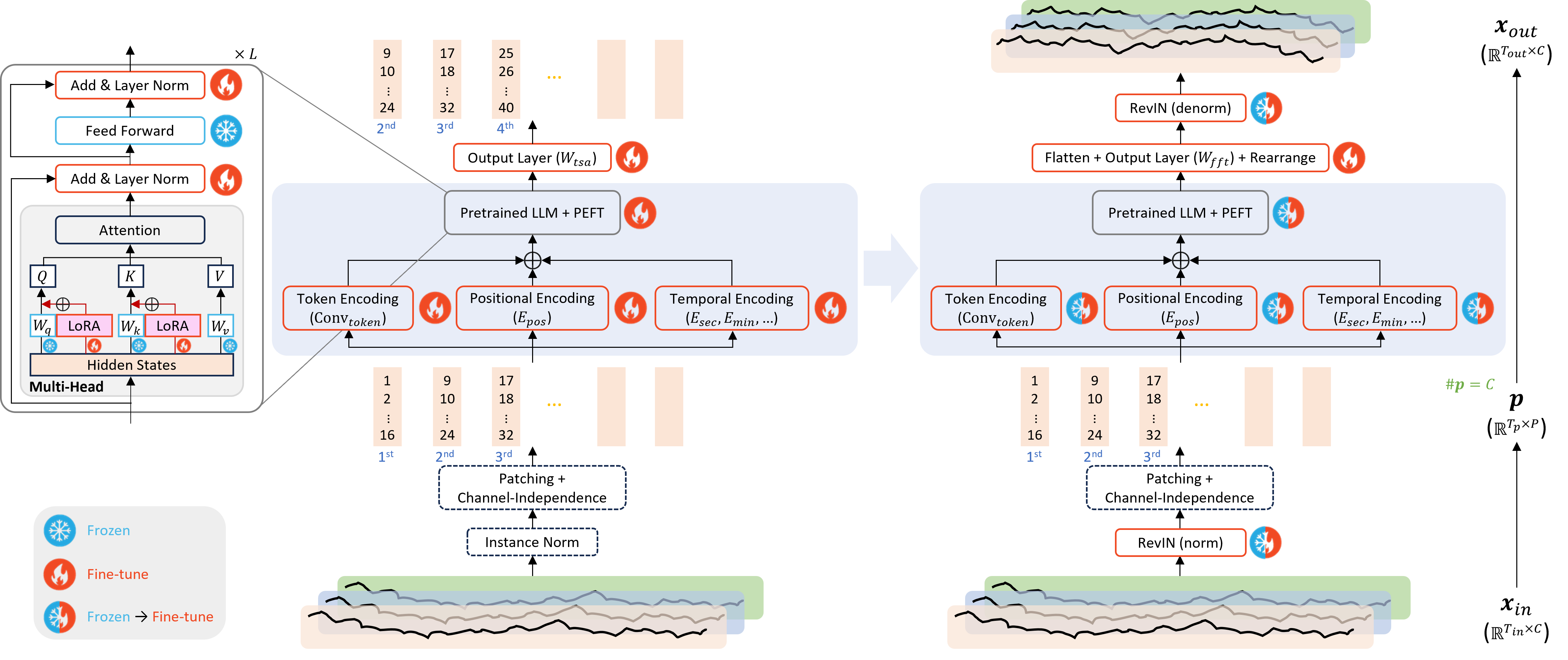}

    \hspace{1cm}
    \begin{subfigure}{.5\textwidth}
        \centering
        \caption{Time-Series Alignment}
        \label{fig:framework_part_a}
    \end{subfigure}
    \hspace{-1cm}
    \begin{subfigure}{.46\textwidth}
        \centering
        \caption{Forecasting Fine-Tuning}
        \label{fig:framework_part_b}
    \end{subfigure}
    
    \caption{\textbf{LLM4TS framework.} The numbers in the patched time-series (e.g., 1, 2, ..., 16 in the first patch) indicate the sequential order of the timestamps. The framework consists of two stages: (a) Time-series alignment, which uses the autoregressive approach to align the pre-trained LLM with patched time-series data.
    (b) Forecasting fine-tuning, which starts with linear probing (i.e., only the output layer is unfrozen), followed by full fine-tuning (all the layers and PEFT components in the LLM are unfrozen).}
    \label{fig:framework}
\end{figure*}

\section{The Proposed LLM4TS}
Fig. \ref{fig:framework} illustrates our LLM4TS framework, leveraging the pre-trained GPT-2 \cite{gpt2} as the backbone model.
We first introduce the time-series alignment stage, which focuses on aligning the LLMs with the characteristics of time-series data using an autoregressive objective (Section \ref{sec:time_series_fine_tuning}).
Subsequently, the forecasting fine-tuning stage is designed to further enhance the model's ability to handle time-series forecasting tasks (Section \ref{sec:forecasting_fine_tuning}).

\subsection{Time-Series Alignment}
\label{sec:time_series_fine_tuning}
Existing LLMs are pre-trained on a general language corpus, which means they fail to learn contextualized information outside linguistic domains; therefore, the \textit{time-series alignment} stage is proposed to align LLMs with the characteristics of time-series data.
Given our selection of GPT-2 \cite{gpt2} as the backbone model, which is a causal language model, we ensure that this stage adopts the same \textit{autoregressive} training methodology used during its pre-training phase.
Fig. \ref{fig:framework}(a) illustrates the autoregressive objective in the time-series alignment stage: given an input sequence of patched time-series data (e.g., \(1^{st}\) patch, \(2^{nd}\) patch, \(3^{rd}\) patch, etc.), the backbone model generates an output sequence shifted one patch to the right (e.g., \(2^{nd}\) patch, \(3^{rd}\) patch, \(4^{th}\) patch, etc.).

\paragraph{Instance Normalization}
Data normalization is essential for stable performance when adapting pre-trained models across various modalities.
Alongside the layer normalization used in the pre-trained LLM, we incorporate instance normalization to improve consistency and reliability in handling diverse time-series datasets.
In our model, instance normalization is employed without incorporating a trainable affine transformation.
This is crucial because when a batch of data is gathered and instance normalization is applied with a trainable affine transformation, the resulting transformed data becomes unsuitable to be the ground truth for the output.
Given that an autoregressive objective is used at this stage, applying a trainable affine transformation is not feasible.

Given an input time-series sample \(\textbf{x}_{in} \in \mathbb{R}^{T_{in} \times C}\), we apply instance normalization (\(\text{IN}\)) to produce a normalized time-series sample \(\textbf{x}_{normed} \in \mathbb{R}^{T_{in} \times C}\) with zero mean and unit standard deviation:
\begin{equation}
    \textbf{x}_{normed} = \text{IN}(\textbf{x}_{in}).
\end{equation}

\paragraph{Time-Series Tokenization}
The context window sizes in pre-trained LLMs (e.g., 1024 in GPT-2) are sufficient for NLP tasks but are inadequate for long-term time-series forecasting.
In our experiments, a prediction length of 720 combined with a look-back window size of 512 easily exceeds these limits.
To address this, we adopt \textit{channel-independence} along with \textit{patching} \cite{patchtst} for time-series tokenization, effectively resolving the context window size constraint and simultaneously reducing the time and space complexity of the Transformer quadratically.
Channel-independence converts multivariate time-series data into multiple univariate time-series data, thus transforming the data's dimension to \(\mathbb{R}^{T_{in} \times 1}\), with the channel dimension \(C\) merged into the batch size dimension.
The subsequent patching step groups adjacent time steps into a singular patch-based token, reducing the input sample's time dimension from \(T_{in}\) to \(T_p\), where \(T_p\) denotes the number of patches, and concurrently expanding the feature dimension from \(1\) to \(P\), with \(P\) representing the patch length.

Given a normalized time-series sample \(\textbf{x}_{normed} \in \mathbb{R}^{T_{in} \times C}\), we first apply channel-independence (\(\text{CI}\)), and then patching to produce a series of patches \( \textbf{p} \in \mathbb{R}^{T_p \times P} \):
\begin{equation}
    \textbf{p} = \text{patching}(\text{CI}(\textbf{x}_{normed})).
\end{equation}

\begin{figure}[t]
    \centering
    \includegraphics[width=1\linewidth]{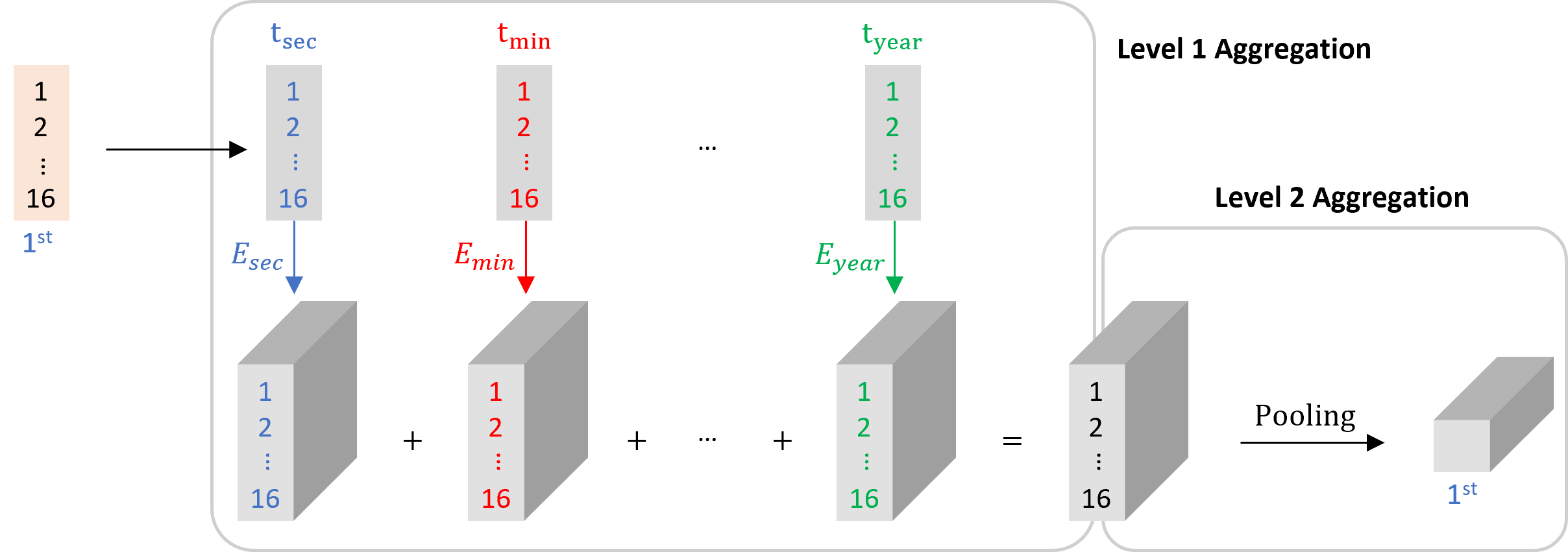}
    \caption{\textbf{Multi-scale temporal encoding for \textit{patched} time-series data}. This process involves a two-level aggregation. Here, only the first patch is shown for simplicity; in practice, all patches in a batch are processed simultaneously. Level 1 aggregation calculates the temporal embedding for each time unit and sums them together. Next, Level 2 aggregation applies a pooling method to extract the final temporal embedding.}
    \label{fig:temporal_encoding}
\end{figure}

\paragraph{Three Encodings for Patched Time-Series Data}
Given our goal to adapt a pre-trained LLM for time-series data, the original token encoding layer (designed for text) becomes unsuitable due to the mismatched modalities.
Additionally, we design a new multi-scale temporal encoding layer to address the inability to process multi-scale temporal information.

Given a series of tokens, applying token encoding is necessary to align their dimensions with the latent embedding dimension of the pre-trained LLM.
In standard NLP practices, this encoding uses a trainable lookup table to map tokens into a high-dimensional space.
However, this method only suits \textit{scalar} tokens, whereas our patched time-series data are \textit{vectors}.
Therefore, we drop the original token encoding layer in the LLM, and employ a one-dimensional convolutional layer \(\text{Conv}_{token}\) as our new token encoding layer.
As opposed to employing a linear layer \cite{gpt4ts}, we choose a convolutional layer due to its superior ability to retain local semantic information within the time-series data.
This results in the generation of the token embedding \(\textbf{e}_{token} \in \mathbb{R}^{T_p \times D}\), where \(D\) denotes the dimension of the embeddings:
\begin{equation}
    \textbf{e}_{token} = \text{Conv}_{token}(\textbf{p}).
\end{equation}

For the positional encoding layer, we employ a trainable lookup table \(E_{pos}\) to map patch locations.
This results in the generation of the positional embedding \(\textbf{e}_{pos} \in \mathbb{R}^{T_p \times D}\):
\begin{equation}
    \textbf{e}_{pos} = E_{pos}(\textbf{i}),
\end{equation}
where \(\textbf{i} \in \mathbb{R}^{T_p}\) represents the indices of the patch locations.

To address the challenge LLMs face in processing multi-scale temporal information, we introduce a multi-scale temporal encoding layer.
When processing time-related data, we face two challenges due to the need to aggregate multiple pieces of information into one unified representation (Fig. \ref{fig:temporal_encoding}):
\begin{enumerate}
    \item Each timestamp includes a range of multi-scale temporal attributes (e.g., seconds, minutes, hours, holidays, etc.).
    \item Each patch encompasses multiple timestamps.
\end{enumerate}
To address the first challenge associated with diverse temporal attributes within a timestamp, we employ Level 1 aggregation: a trainable lookup table for each temporal attribute (e.g.,  \(E_{sec},E_{min},...\)), mapping it into a high-dimensional space, and then summing them to produce a singular temporal embedding.
In response to the second challenge of multiple timestamps within a patch, we use Level 2 aggregation: a pooling method to extract the final temporal embedding.
For the pooling method, we opt for the ``select first" method, where the initial timestamp is designated as representative of the entire patch.
This is because the first timestamp often carries the most significant and representative information for the entire duration covered by the patch, especially in time-series data where earlier events can have a substantial influence on the subsequent sequence.
This process generates the final temporal embedding \(\textbf{e}_{temp} \in \mathbb{R}^{T_p \times D}\):
\begin{equation}
    \textbf{e}_{temp} = \text{Pooling}\left( \sum_{a \in \{\text{sec}, \text{min}, \text{hour}, \ldots\}} E_{a}(\textbf{t}_{a}) \right),
\end{equation}
where \( a \) represents different temporal attributes (seconds, minutes, hours, holidays, etc.), \( E_{a} \) denotes the trainable lookup table for each temporal attribute, \( \textbf{t}_{a} \in \mathbb{R}^{T_{p} \times P} \) are the series of patches containing temporal information for that temporal attribute, and \(\text{Pooling}\) applies the pooling method to the aggregated embeddings.

Finally, the token, positional, and temporal embeddings are summed to yield the final embedding \(\textbf{e} \in \mathbb{R}^{T_p \times D}\), which is then fed into the pre-trained Transformer blocks:
\begin{equation}
    \textbf{e} = \textbf{e}_{token} + \textbf{e}_{pos} + \textbf{e}_{temp}.
\end{equation}

\paragraph{Pre-Trained LLM}
To preserve LLMs' data-independent representation learning capability, most parameters in these LLMs are kept fixed.
Empirical evidence \cite{fpt,gpt4ts} shows that training these LLMs from scratch often hurts performance, highlighting the importance of fixing most parameters to retain the LLM's representation learning capability.
To that end, we opt for freezing most parameters, particularly those associated with the multi-head attention and feed-forward layers in the Transformer block, as they are the most responsible for representation learning \cite{gpt4ts}.

For the remaining trainable parameters in the pre-trained LLM, we employ two Parameter-Efficient Fine-Tuning (PEFT) methods to selectively adjust or introduce a limited set of trainable parameters.
Specifically, we utilize Layer Normalization Tuning \cite{fpt} to adjust pre-existing parameters in Transformer blocks, making the affine transformation in layer normalization trainable.
Concurrently, we employ Low-Rank Adaptation (LoRA) \cite{lora}, which introduces trainable low-rank matrices that are applied to the query (\(\text{Q}\)) and key (\(\text{K}\)) matrices in the self-attention mechanism.
With these two PEFT techniques, only \(1.5\%\) of the pre-trained LLM's total parameters are used to be trained.

Given the embedding \(\textbf{e}\) (which is adjusted to the required embedding dimension \(D\) by three encoding layers), we pass it into the pre-trained LLM, which comprises a series of pre-trained Transformer blocks (\(\text{TBs}\)) (with \(L\) blocks in total).
This process yields the final embeddings \( \textbf{z} \in \mathbb{R}^{T_p \times D} \):
\begin{equation}
    \textbf{z} = \text{TBs}(\textbf{e}).
\end{equation}

After being processed by the pre-trained LLM, we employ a linear output layer \( W_{tsa} \in \mathbb{R}^{P \times D} \) to transform the output embedding back to patched time-series data:
\begin{equation}
    \hat{\textbf{p}}_{shifted} = \textbf{z}W_{tsa}^\top,
\end{equation}
where \(\hat{\textbf{p}}_{shifted} \in \mathbb{R}^{T_p \times P}\) represents our prediction target, corresponding to the original time-series patches (\(\textbf{p}\)) shifted one patch to the right, in line with the autoregressive objective of this stage.
To ensure the prediction precisely reconstructs the actual shifted patched data \( \textbf{p}_{shifted} \in \mathbb{R}^{T_p \times P}\), we use the Mean Squared Error (MSE) as the loss function:
\begin{equation}
\mathcal{L}_{tsa} = \text{MSE} (\textbf{p}_{shifted}, \hat{\textbf{p}}_{shifted}).
\end{equation}

\subsection{Forecasting Fine-tuning}
\label{sec:forecasting_fine_tuning}
After aligning the pre-trained LLM with patched time-series data in the time-series alignment stage, we transfer the trained weights of the backbone model, including those from the encoding layers, to the \textit{forecasting fine-tuning} stage.
When fine-tuning the backbone model for the forecasting task, two primary training strategies are available: full fine-tuning (where all model parameters are updated) and linear probing (where only the final linear output layer is updated). 
Studies have shown that a sequential approach—initial linear probing followed by full fine-tuning (LP-FT, as illustrated in Fig. \ref{fig:framework}(b))—consistently surpasses strategies exclusively employing either method \cite{lp_ft}.
The superiority of LP-FT is due to its dual-phase approach: first finding an optimized output layer to minimize later adjustments in fine-tuning (preserving feature extractor efficacy for out-of-distribution (OOD) scenarios), and then employing full fine-tuning to adapt the model to the specific task (enhancing in-distribution (ID) accuracy) \cite{lp_ft}.

For the model architecture in the forecasting fine-tuning stage, we preserve most of the structure as in the time-series alignment stage, including the three encoding layers and the pre-trained LLM.
However, there are two architectural differences in this stage: instance normalization and the output layer.

The first architectural difference is in the instance normalization, where we adopt Reversible Instance Normalization (RevIN) \cite{revin} to enhance forecasting accuracy.
RevIN involves batch-specific instance normalization and subsequent denormalization, both sharing the same trainable affine transformation.
The additional denormalization step addresses distribution shifts between training and testing data, which is a common challenge in the time-series domain (e.g., seasonal changes).
Therefore, during the time-series tokenization step, we apply RevIN's normalization, succeeded by channel-independence and patching:
\begin{equation}
    \textbf{p} = \text{patching}(\text{CI}(\text{RevIN}_{norm}(\textbf{x}_{in}))).
\end{equation}
Notably, the denormalization step is applicable only to unpatched time-series data; hence, in the time-series alignment stage, standard instance normalization is employed.

The second architectural difference lies in the output layer, whose function is to transform the final embedding \(\textbf{z}\) into the predicted future data, presented in the general (unpatched) time-series format.
This involves flattening the data and passing it through the linear output layer \( W_{fft} \in \mathbb{R}^{T_{out} \times T_{p} \cdot D} \), followed by rearrangement, and then applying RevIN's denormalization to obtain the final prediction \(\hat{\textbf{x}}_{out} \in \mathbb{R}^{T_{out} \times C}\):
\begin{equation}
    \hat{\textbf{x}}_{out} = \text{RevIN}_{denorm}(\text{Rearrange}((\text{Flatten}(\textbf{z}))W_{fft}^\top).
\end{equation}
To ensure that this prediction accurately reconstructs the future data \( \textbf{x}_{out} \in \mathbb{R}^{T_{out} \times C}\), we use MSE as the loss function:
\begin{equation}
\mathcal{L}_{fft} = \text{MSE} (\textbf{x}_{out}, \hat{\textbf{x}}_{out}).
\end{equation}

\renewcommand{\arraystretch}{1.2}  
\setlength{\tabcolsep}{24pt}       

\begin{table}[]
\centering
\caption{\textbf{Statistical overview of the 7 datasets for long-term time-series forecasting.}}
\label{tab:dataset}
\begin{tabular}{c|ccc}
\hline
Datasets & Features & Timesteps & Granularity \\ \hline
Weather & 21 & 52,696 & 10 min \\
Traffic & 862 & 17,544 & 1 hour \\
Electricity & 321 & 26,304 & 1 hour \\
ETTh1 \& ETTh2 & 7 & 17,420 & 1 hour \\
ETTm1 \& ETTm2 & 7 & 69,680 & 5 min \\ \hline
\end{tabular}
\end{table}

\section{Experiments}
\paragraph{\textbf{Datasets}}
In our long-term forecasting analysis, we experiment on 7 real-world, publicly accessible benchmark datasets.
We present detailed statistics for these datasets in Table \ref{tab:dataset}, including the number of features, the total length of the datasets, and their sampling frequency.

\textbf{Weather}\footnote{https://www.ncei.noaa.gov/data/local-climatological-data/} comprises local climatological data spanning 4 years for approximately 1,600 locations in the U.S.
It includes 11 weather variables in each record, in addition to the target variable 'wet bulb.'
\textbf{Traffic}\footnote{http://pems.dot.ca.gov/} consists of hourly observations from the California Department of Transportation, detailing road occupancy rates captured by various sensors located on freeways in the San Francisco Bay area.
\textbf{Electricity}\footnote{https://archive.ics.uci.edu/dataset/321/electricityloaddiagrams20112014} comprises hourly power usage data for 321 customers, spanning from 2012 to 2014, with 'MT\_320' set as the target variable.
\textbf{ETT}  \cite{informer} focuses on long-duration electric power deployment data.
The collection includes two datasets sampled hourly (ETTh1, ETTh2) and two datasets sampled every 15 minutes (ETTm1, ETTm2), covering a period of over two years from various provinces in China.
Each dataset in the ETT series features one oil temperature variable alongside six power load variables.

\paragraph{\textbf{Evaluation Metrics}}
In time-series forecasting, Mean Squared Error (MSE) and Mean Absolute Error (MAE) are commonly utilized metrics for evaluating performance.
The Mean Squared Error (MSE) can be expressed as:
\begin{equation}
\text{MSE} = \frac{1}{N} \sum_{n=1}^{N} (x_{\text{out}} - \hat{x}_{\text{out}})^2.
\end{equation}
Here, \(x_{\text{out}}\) indicates the actual future data that corresponds to the past data \(x_{\text{in}}\), while \(\hat{x}_{\text{out}}\) represents the predicted future data based on the input past data.
\(N\) is the total number of samples.
The Mean Absolute Error (MAE) is given by:
\begin{equation}
\text{MAE} = \frac{1}{N} \sum_{n=1}^{N} |x_{\text{out}} - \hat{x}_{\text{out}}|.
\end{equation}

\paragraph{\textbf{Baselines}}
For long-term time-series forecasting, we focus on a range of state-of-the-art models.
The same set of models is used for few-shot learning and ablation studies.
\textbf{GPT4TS} \cite{gpt4ts} employs patching and channel independence to initially transform time-series data into tokens.
It then utilizes a pre-trained Large Language Model (GPT-2), maintaining the pre-trained weights in the self-attention and feedforward layers of the residual blocks from the pre-trained LLM.
\textbf{DLinear} \cite{dlinear} challenges the prevailing use of Transformer-based models for long-term time-series forecasting.
It introduces a simple one-layer linear model that surprisingly outperforms complex Transformer-based models on several real datasets.
\textbf{PatchTST} \cite{patchtst} introduces a Transformer-based approach for time-series forecasting, focusing on efficiency by employing patching and channel-independence to transform time-series data into patches.
\textbf{FEDformer} \cite{fedformer} combines Transformers with seasonal-trend decomposition and frequency enhancement for efficient and effective long-term series forecasting, overcoming traditional Transformer limitations by capturing both global trends and detailed structures with linear complexity.
\redtext{
\textbf{Time-LLM} \cite{time_llm} reprograms LLMs for time series forecasting by converting time series into text prototypes and using prompts to guide prediction. It outperforms specialized models, particularly in few-shot and zero-shot settings.
\textbf{TEMPO} \cite{tempo} adapts GPT-like models for time series forecasting by decomposing trends and using prompts for better distribution adaptation, excelling in zero-shot scenarios.
\textbf{TEST} \cite{test} aligns time series with LLM embeddings through tokenization and contrastive learning, enabling effective time series forecasting using pre-trained LLMs without fine-tuning.
}

For unsupervised representation learning in time-series analysis, we explore advanced models that excel in extracting meaningful representations without relying on labeled data.
\textbf{PatchTST} \cite{patchtst} in this context is a variant distinct from the one used in forecasting experiments, with an emphasis on representation learning.
It adopts an MLM (Masked Language Model) strategy, similar to BERT, to learn representations.
\textbf{BTSF} \cite{btsf} introduces a Bilinear Temporal-Spectral Fusion framework to improve time-series representation learning by integrating temporal and spectral information.
This approach minimizes sampling bias and optimizes feature representation through instance-level augmentation and fusion techniques.
\textbf{TS2Vec} \cite{ts2vec} is the first universal framework dedicated to learning representations of time-series data.
It emphasizes distinguishing multi-scale contextual information on both the instance and timestamp levels, demonstrating effectiveness in various time-series tasks.
\textbf{TNC} \cite{tnc} utilizes the Augmented Dickey-Fuller test to detect temporal neighborhoods and implements Positive-Unlabeled learning to address sampling bias.
\textbf{TS-TCC} \cite{ts_tcc} produces two different views via strong and weak augmentations and enhances representations through contrastive learning, focusing on the temporal and contextual differences between these views.

\renewcommand{\arraystretch}{1.06}  
\setlength{\tabcolsep}{1.2pt}       

\begin{table}[h]
\centering
\small
\caption{\redtext{\textbf{Few-shot long-term forecasting using \(5\%\) of the training data.} For most datasets, results are reported over prediction lengths \(T_{out} \in \{96,192,336,720\}\). However, for datasets marked with * (ETTh1, ETTh2, and Traffic), only \(T_{out} \in \{96,192,336\}\) are used because there are insufficient data to constitute a training set when \(T_{out} = 720\). The best results are in {\color[HTML]{FF0000} \textbf{bold}}, while the second-best results are in {\color[HTML]{0000FF} {\ul underlined}}. Note that TEMPO* is evaluated under a zero-shot setting as it is a prompt-based method.}}
\label{tab:few_shot_learning_5}
\begin{tabular}{cc|cc|cc|cc|cc|cc|cc|cc}
\hline
\multicolumn{2}{c|}{Methods} & \multicolumn{2}{c|}{LLM4TS} & \multicolumn{2}{c|}{GPT4TS} & \multicolumn{2}{c|}{DLinear} & \multicolumn{2}{c|}{PatchTST} & \multicolumn{2}{c|}{Time-LLM} & \multicolumn{2}{c|}{TEMPO*} & \multicolumn{2}{c}{TEST} \\ \hline
\multicolumn{2}{c|}{Metric} & MSE & MAE & MSE & MAE & MSE & MAE & MSE & MAE & MSE & MAE & MSE & MAE & MSE & MAE \\ \hline
\multicolumn{1}{c|}{} & 96 & 0.173 & {\color[HTML]{0000FF} {\ul 0.227}} & 0.175 & 0.230 & 0.184 & 0.242 & {\color[HTML]{FF0000} \textbf{0.171}} & {\color[HTML]{FF0000} \textbf{0.224}} & {\color[HTML]{0000FF} {\ul 0.172}} & 0.263 & 0.211 & 0.254 & 0.182 & 0.276 \\
\multicolumn{1}{c|}{} & 192 & {\color[HTML]{FF0000} \textbf{0.218}} & {\color[HTML]{FF0000} \textbf{0.265}} & 0.227 & 0.276 & 0.228 & 0.283 & 0.230 & 0.277 & {\color[HTML]{0000FF} {\ul 0.224}} & {\color[HTML]{0000FF} {\ul 0.271}} & 0.254 & 0.298 & 0.273 & 0.283 \\
\multicolumn{1}{c|}{} & 336 & {\color[HTML]{FF0000} \textbf{0.276}} & {\color[HTML]{FF0000} \textbf{0.310}} & 0.286 & 0.322 & {\color[HTML]{0000FF} {\ul 0.279}} & 0.322 & 0.294 & 0.326 & 0.282 & {\color[HTML]{0000FF} {\ul 0.321}} & 0.292 & 0.332 & 0.294 & 0.325 \\
\multicolumn{1}{c|}{\multirow{-4}{*}{Weather}} & 720 & {\color[HTML]{FF0000} \textbf{0.355}} & {\color[HTML]{FF0000} \textbf{0.366}} & 0.366 & {\color[HTML]{0000FF} {\ul 0.379}} & {\color[HTML]{0000FF} {\ul 0.364}} & 0.388 & 0.384 & 0.387 & 0.366 & 0.381 & 0.370 & {\color[HTML]{0000FF} {\ul 0.379}} & 0.383 & 0.388 \\ \hline
\multicolumn{1}{c|}{} & 96 & 0.509 & 0.484 & 0.543 & 0.506 & 0.547 & 0.503 & 0.557 & 0.519 & {\color[HTML]{0000FF} {\ul 0.483}} & 0.464 & {\color[HTML]{FF0000} \textbf{0.400}} & {\color[HTML]{FF0000} \textbf{0.406}} & 0.531 & {\color[HTML]{0000FF} {\ul 0.447}} \\
\multicolumn{1}{c|}{} & 192 & 0.717 & 0.581 & 0.748 & 0.580 & 0.720 & 0.604 & 0.711 & 0.570 & {\color[HTML]{0000FF} {\ul 0.629}} & 0.540 & {\color[HTML]{FF0000} \textbf{0.426}} & {\color[HTML]{FF0000} \textbf{0.421}} & 0.750 & {\color[HTML]{0000FF} {\ul 0.533}} \\
\multicolumn{1}{c|}{} & 336 & {\color[HTML]{0000FF} {\ul 0.728}} & {\color[HTML]{0000FF} {\ul 0.589}} & 0.754 & 0.595 & 0.984 & 0.727 & 0.816 & 0.619 & 0.768 & 0.626 & {\color[HTML]{FF0000} \textbf{0.441}} & {\color[HTML]{FF0000} \textbf{0.430}} & 0.741 & 0.636 \\
\multicolumn{1}{c|}{\multirow{-4}{*}{ETTh1}} & 720 & - & - & - & - & - & - & - & - & - & - & - & - & - & - \\ \hline
\multicolumn{1}{c|}{} & 96 & {\color[HTML]{0000FF} {\ul 0.314}} & {\color[HTML]{0000FF} {\ul 0.375}} & 0.376 & 0.421 & 0.442 & 0.456 & 0.401 & 0.421 & 0.336 & 0.397 & {\color[HTML]{FF0000} \textbf{0.301}} & {\color[HTML]{FF0000} \textbf{0.353}} & 0.368 & 0.457 \\
\multicolumn{1}{c|}{} & 192 & {\color[HTML]{0000FF} {\ul 0.365}} & {\color[HTML]{0000FF} {\ul 0.408}} & 0.418 & 0.441 & 0.617 & 0.542 & 0.452 & 0.455 & 0.406 & 0.425 & {\color[HTML]{FF0000} \textbf{0.355}} & {\color[HTML]{FF0000} \textbf{0.389}} & 0.407 & 0.486 \\
\multicolumn{1}{c|}{} & 336 & {\color[HTML]{0000FF} {\ul 0.398}} & 0.432 & 0.408 & 0.439 & 1.424 & 0.849 & 0.464 & 0.469 & 0.405 & 0.432 & {\color[HTML]{FF0000} \textbf{0.379}} & {\color[HTML]{FF0000} \textbf{0.408}} & 0.402 & {\color[HTML]{0000FF} {\ul 0.428}} \\
\multicolumn{1}{c|}{\multirow{-4}{*}{ETTh2}} & 720 & - & - & - & - & - & - & - & - & - & - & - & - & - & - \\ \hline
\multicolumn{1}{c|}{} & 96 & 0.349 & 0.379 & 0.386 & 0.405 & {\color[HTML]{0000FF} {\ul 0.332}} & {\color[HTML]{FF0000} \textbf{0.374}} & 0.399 & 0.414 & {\color[HTML]{FF0000} \textbf{0.316}} & {\color[HTML]{0000FF} {\ul 0.377}} & 0.438 & 0.424 & 0.340 & 0.381 \\
\multicolumn{1}{c|}{} & 192 & {\color[HTML]{0000FF} {\ul 0.374}} & {\color[HTML]{0000FF} {\ul 0.394}} & 0.440 & 0.438 & {\color[HTML]{FF0000} \textbf{0.358}} & {\color[HTML]{FF0000} \textbf{0.390}} & 0.441 & 0.436 & 0.450 & 0.464 & 0.461 & 0.432 & 0.473 & 0.451 \\
\multicolumn{1}{c|}{} & 336 & {\color[HTML]{0000FF} {\ul 0.411}} & {\color[HTML]{0000FF} {\ul 0.417}} & 0.485 & 0.459 & {\color[HTML]{FF0000} \textbf{0.402}} & {\color[HTML]{FF0000} \textbf{0.416}} & 0.499 & 0.467 & 0.450 & 0.424 & 0.515 & 0.467 & 0.519 & 0.464 \\
\multicolumn{1}{c|}{\multirow{-4}{*}{ETTm1}} & 720 & 0.516 & {\color[HTML]{0000FF} {\ul 0.479}} & 0.577 & 0.499 & {\color[HTML]{0000FF} {\ul 0.511}} & 0.489 & 0.767 & 0.587 & {\color[HTML]{FF0000} \textbf{0.483}} & {\color[HTML]{FF0000} \textbf{0.471}} & 0.591 & 0.509 & 0.604 & 0.499 \\ \hline
\multicolumn{1}{c|}{} & 96 & 0.192 & 0.273 & 0.199 & 0.280 & 0.236 & 0.326 & 0.206 & 0.288 & {\color[HTML]{FF0000} \textbf{0.174}} & {\color[HTML]{FF0000} \textbf{0.261}} & {\color[HTML]{0000FF} {\ul 0.185}} & {\color[HTML]{0000FF} {\ul 0.267}} & 0.254 & 0.275 \\
\multicolumn{1}{c|}{} & 192 & 0.249 & 0.309 & 0.256 & 0.316 & 0.306 & 0.373 & 0.264 & 0.324 & {\color[HTML]{FF0000} \textbf{0.215}} & {\color[HTML]{0000FF} {\ul 0.287}} & {\color[HTML]{0000FF} {\ul 0.243}} & 0.304 & 0.265 & {\color[HTML]{FF0000} \textbf{0.286}} \\
\multicolumn{1}{c|}{} & 336 & {\color[HTML]{0000FF} {\ul 0.301}} & {\color[HTML]{0000FF} {\ul 0.342}} & 0.318 & 0.353 & 0.380 & 0.423 & 0.334 & 0.367 & {\color[HTML]{FF0000} \textbf{0.273}} & {\color[HTML]{FF0000} \textbf{0.330}} & 0.309 & 0.345 & 0.360 & 0.373 \\
\multicolumn{1}{c|}{\multirow{-4}{*}{ETTm2}} & 720 & {\color[HTML]{0000FF} {\ul 0.402}} & {\color[HTML]{0000FF} {\ul 0.405}} & 0.460 & 0.436 & 0.674 & 0.583 & 0.454 & 0.432 & 0.433 & 0.412 & {\color[HTML]{FF0000} \textbf{0.386}} & {\color[HTML]{FF0000} \textbf{0.395}} & 0.511 & 0.439 \\ \hline
\multicolumn{1}{c|}{} & 96 & {\color[HTML]{FF0000} \textbf{0.139}} & {\color[HTML]{FF0000} \textbf{0.235}} & {\color[HTML]{0000FF} {\ul 0.143}} & {\color[HTML]{0000FF} {\ul 0.241}} & 0.150 & 0.251 & 0.145 & 0.244 & 0.147 & 0.242 & 0.178 & 0.276 & 0.144 & 0.246 \\
\multicolumn{1}{c|}{} & 192 & {\color[HTML]{FF0000} \textbf{0.155}} & 0.249 & 0.159 & 0.255 & 0.163 & 0.263 & 0.163 & 0.260 & {\color[HTML]{0000FF} {\ul 0.158}} & {\color[HTML]{FF0000} \textbf{0.241}} & 0.198 & 0.293 & 0.180 & {\color[HTML]{0000FF} {\ul 0.248}} \\
\multicolumn{1}{c|}{} & 336 & {\color[HTML]{FF0000} \textbf{0.174}} & {\color[HTML]{FF0000} \textbf{0.269}} & 0.179 & {\color[HTML]{0000FF} {\ul 0.274}} & {\color[HTML]{0000FF} {\ul 0.175}} & 0.278 & 0.183 & 0.281 & 0.178 & 0.277 & 0.209 & 0.309 & 0.194 & 0.304 \\
\multicolumn{1}{c|}{\multirow{-4}{*}{ECL}} & 720 & 0.222 & {\color[HTML]{0000FF} {\ul 0.310}} & 0.233 & 0.323 & {\color[HTML]{0000FF} {\ul 0.219}} & 0.311 & 0.233 & 0.323 & 0.224 & 0.312 & 0.279 & 0.355 & {\color[HTML]{FF0000} \textbf{0.205}} & {\color[HTML]{FF0000} \textbf{0.277}} \\ \hline
\multicolumn{1}{c|}{} & 96 & {\color[HTML]{FF0000} \textbf{0.401}} & {\color[HTML]{FF0000} \textbf{0.285}} & 0.419 & 0.298 & 0.427 & 0.304 & {\color[HTML]{0000FF} {\ul 0.404}} & {\color[HTML]{0000FF} {\ul 0.286}} & 0.414 & 0.291 & 0.476 & 0.343 & 0.443 & 0.317 \\
\multicolumn{1}{c|}{} & 192 & 0.418 & {\color[HTML]{0000FF} {\ul 0.293}} & 0.434 & 0.305 & 0.447 & 0.315 & {\color[HTML]{0000FF} {\ul 0.412}} & 0.294 & 0.419 & {\color[HTML]{FF0000} \textbf{0.291}} & 0.496 & 0.355 & {\color[HTML]{FF0000} \textbf{0.407}} & 0.320 \\
\multicolumn{1}{c|}{} & 336 & {\color[HTML]{FF0000} \textbf{0.436}} & {\color[HTML]{FF0000} \textbf{0.308}} & 0.449 & 0.313 & 0.478 & 0.333 & 0.439 & {\color[HTML]{0000FF} {\ul 0.310}} & {\color[HTML]{0000FF} {\ul 0.437}} & 0.314 & 0.503 & 0.356 & 0.440 & 0.323 \\
\multicolumn{1}{c|}{\multirow{-4}{*}{Traffic}} & 720 & - & - & - & - & - & - & - & - & - & - & - & - & - & - \\ \hline
\multicolumn{2}{c|}{Avg. Rank} & {\color[HTML]{FF0000} \textbf{2.120}} & {\color[HTML]{FF0000} \textbf{2.200}} & 4.200 & 4.000 & 4.680 & 5.080 & 4.760 & 4.640 & {\color[HTML]{0000FF} {\ul 2.720}} & {\color[HTML]{0000FF} {\ul 2.920}} & 4.400 & 4.280 & 5.000 & 4.640 \\ \hline
\end{tabular}
\end{table}
\renewcommand{\arraystretch}{1.1}  
\setlength{\tabcolsep}{1.5pt}       

\begin{table}[h]
\centering
\small
\caption{\redtext{\textbf{Few-shot long-term forecasting using \(10\%\) of the training data.} We use prediction lengths \(T \in \{96,192,336,720\}\) for all datasets. The best results are in {\color[HTML]{FF0000} \textbf{bold}}, while the second-best results are {\color[HTML]{0000FF} {\ul underlined}}. Note that TEMPO* is evaluated under a zero-shot setting as it is a prompt-based method.}}
\label{tab:few_shot_learning_10}
\begin{tabular}{cc|cc|cc|cc|cc|cc|cc|cc}
\hline
\multicolumn{2}{c|}{Methods} & \multicolumn{2}{c|}{LLM4TS} & \multicolumn{2}{c|}{GPT4TS} & \multicolumn{2}{c|}{DLinear} & \multicolumn{2}{c|}{PatchTST} & \multicolumn{2}{c|}{Time-LLM} & \multicolumn{2}{c|}{TEMPO*} & \multicolumn{2}{c}{TEST} \\ \hline
\multicolumn{2}{c|}{Metric} & MSE & MAE & MSE & MAE & MSE & MAE & MSE & MAE & MSE & MAE & MSE & MAE & MSE & MAE \\ \hline
\multicolumn{1}{c|}{} & 96 & {\color[HTML]{FF0000} \textbf{0.158}} & {\color[HTML]{FF0000} \textbf{0.207}} & 0.163 & 0.215 & 0.171 & 0.224 & 0.165 & 0.215 & {\color[HTML]{0000FF} {\ul 0.161}} & {\color[HTML]{0000FF} {\ul 0.210}} & 0.211 & 0.254 & 0.163 & 0.213 \\
\multicolumn{1}{c|}{} & 192 & {\color[HTML]{FF0000} \textbf{0.204}} & {\color[HTML]{0000FF} {\ul 0.249}} & 0.210 & 0.254 & 0.215 & 0.263 & 0.210 & 0.257 & {\color[HTML]{FF0000} \textbf{0.204}} & {\color[HTML]{FF0000} \textbf{0.248}} & 0.254 & 0.298 & 0.230 & 0.263 \\
\multicolumn{1}{c|}{} & 336 & {\color[HTML]{FF0000} \textbf{0.254}} & {\color[HTML]{0000FF} {\ul 0.288}} & {\color[HTML]{0000FF} {\ul 0.256}} & 0.292 & 0.258 & 0.299 & 0.259 & 0.297 & 0.261 & 0.302 & 0.292 & 0.332 & 0.258 & {\color[HTML]{FF0000} \textbf{0.282}} \\
\multicolumn{1}{c|}{\multirow{-4}{*}{Weather}} & 720 & 0.322 & 0.336 & 0.321 & 0.339 & 0.320 & 0.346 & 0.332 & 0.346 & {\color[HTML]{0000FF} {\ul 0.309}} & {\color[HTML]{0000FF} {\ul 0.332}} & 0.370 & 0.379 & {\color[HTML]{FF0000} \textbf{0.301}} & {\color[HTML]{FF0000} \textbf{0.328}} \\ \hline
\multicolumn{1}{c|}{} & 96 & {\color[HTML]{0000FF} {\ul 0.417}} & {\color[HTML]{0000FF} {\ul 0.432}} & 0.458 & 0.456 & 0.492 & 0.495 & 0.516 & 0.485 & 0.448 & 0.460 & {\color[HTML]{FF0000} \textbf{0.400}} & {\color[HTML]{FF0000} \textbf{0.406}} & 0.455 & 0.457 \\
\multicolumn{1}{c|}{} & 192 & {\color[HTML]{0000FF} {\ul 0.469}} & {\color[HTML]{0000FF} {\ul 0.468}} & 0.570 & 0.516 & 0.565 & 0.538 & 0.598 & 0.524 & 0.484 & 0.483 & {\color[HTML]{FF0000} \textbf{0.426}} & {\color[HTML]{FF0000} \textbf{0.421}} & 0.572 & 0.519 \\
\multicolumn{1}{c|}{} & 336 & {\color[HTML]{0000FF} {\ul 0.505}} & {\color[HTML]{0000FF} {\ul 0.499}} & 0.608 & 0.535 & 0.721 & 0.622 & 0.657 & 0.550 & 0.589 & 0.540 & {\color[HTML]{FF0000} \textbf{0.441}} & {\color[HTML]{FF0000} \textbf{0.430}} & 0.578 & 0.531 \\
\multicolumn{1}{c|}{\multirow{-4}{*}{ETTh1}} & 720 & 0.708 & {\color[HTML]{0000FF} {\ul 0.572}} & 0.725 & 0.591 & 0.986 & 0.743 & 0.762 & 0.610 & {\color[HTML]{0000FF} {\ul 0.700}} & 0.604 & {\color[HTML]{FF0000} \textbf{0.443}} & {\color[HTML]{FF0000} \textbf{0.451}} & 0.723 & 0.594 \\ \hline
\multicolumn{1}{c|}{} & 96 & {\color[HTML]{0000FF} {\ul 0.282}} & {\color[HTML]{0000FF} {\ul 0.351}} & 0.331 & 0.374 & 0.357 & 0.411 & 0.353 & 0.389 & {\color[HTML]{FF0000} \textbf{0.275}} & {\color[HTML]{FF0000} \textbf{0.326}} & 0.301 & 0.353 & 0.332 & 0.374 \\
\multicolumn{1}{c|}{} & 192 & {\color[HTML]{0000FF} {\ul 0.364}} & 0.400 & 0.402 & 0.411 & 0.569 & 0.519 & 0.403 & 0.414 & 0.374 & {\color[HTML]{FF0000} \textbf{0.373}} & {\color[HTML]{FF0000} \textbf{0.355}} & {\color[HTML]{0000FF} {\ul 0.389}} & 0.401 & 0.433 \\
\multicolumn{1}{c|}{} & 336 & {\color[HTML]{FF0000} \textbf{0.374}} & {\color[HTML]{0000FF} {\ul 0.416}} & 0.406 & 0.433 & 0.671 & 0.572 & 0.426 & 0.441 & 0.406 & 0.429 & {\color[HTML]{0000FF} {\ul 0.379}} & {\color[HTML]{FF0000} \textbf{0.408}} & 0.408 & 0.440 \\
\multicolumn{1}{c|}{\multirow{-4}{*}{ETTh2}} & 720 & 0.445 & 0.461 & 0.449 & 0.464 & 0.824 & 0.648 & 0.477 & 0.480 & {\color[HTML]{0000FF} {\ul 0.427}} & {\color[HTML]{0000FF} {\ul 0.449}} & {\color[HTML]{FF0000} \textbf{0.409}} & {\color[HTML]{FF0000} \textbf{0.440}} & 0.459 & 0.480 \\ \hline
\multicolumn{1}{c|}{} & 96 & 0.360 & {\color[HTML]{FF0000} \textbf{0.388}} & 0.390 & 0.404 & {\color[HTML]{0000FF} {\ul 0.352}} & 0.392 & 0.410 & 0.419 & {\color[HTML]{FF0000} \textbf{0.346}} & {\color[HTML]{FF0000} \textbf{0.388}} & 0.438 & 0.424 & 0.392 & 0.401 \\
\multicolumn{1}{c|}{} & 192 & 0.386 & {\color[HTML]{FF0000} \textbf{0.401}} & 0.429 & 0.423 & {\color[HTML]{0000FF} {\ul 0.382}} & {\color[HTML]{0000FF} {\ul 0.412}} & 0.437 & 0.434 & {\color[HTML]{FF0000} \textbf{0.373}} & 0.416 & 0.461 & 0.432 & 0.423 & 0.426 \\
\multicolumn{1}{c|}{} & 336 & {\color[HTML]{0000FF} {\ul 0.415}} & {\color[HTML]{FF0000} \textbf{0.417}} & 0.469 & 0.439 & 0.419 & 0.434 & 0.476 & 0.454 & {\color[HTML]{FF0000} \textbf{0.413}} & {\color[HTML]{0000FF} {\ul 0.426}} & 0.515 & 0.467 & 0.471 & 0.444 \\
\multicolumn{1}{c|}{\multirow{-4}{*}{ETTm1}} & 720 & {\color[HTML]{FF0000} \textbf{0.470}} & {\color[HTML]{FF0000} \textbf{0.445}} & 0.569 & 0.498 & 0.490 & 0.477 & 0.681 & 0.556 & {\color[HTML]{0000FF} {\ul 0.485}} & {\color[HTML]{0000FF} {\ul 0.476}} & 0.591 & 0.509 & 0.552 & 0.501 \\ \hline
\multicolumn{1}{c|}{} & 96 & {\color[HTML]{0000FF} {\ul 0.184}} & 0.265 & 0.188 & 0.269 & 0.213 & 0.303 & 0.191 & 0.274 & {\color[HTML]{FF0000} \textbf{0.177}} & {\color[HTML]{FF0000} \textbf{0.261}} & 0.185 & 0.267 & 0.233 & {\color[HTML]{0000FF} {\ul 0.262}} \\
\multicolumn{1}{c|}{} & 192 & {\color[HTML]{FF0000} \textbf{0.240}} & {\color[HTML]{FF0000} \textbf{0.301}} & 0.251 & 0.309 & 0.278 & 0.345 & 0.252 & 0.317 & {\color[HTML]{0000FF} {\ul 0.241}} & 0.314 & 0.243 & 0.304 & 0.303 & {\color[HTML]{0000FF} {\ul 0.302}} \\
\multicolumn{1}{c|}{} & 336 & {\color[HTML]{0000FF} {\ul 0.294}} & {\color[HTML]{0000FF} {\ul 0.337}} & 0.307 & 0.346 & 0.338 & 0.385 & 0.306 & 0.353 & {\color[HTML]{FF0000} \textbf{0.274}} & {\color[HTML]{FF0000} \textbf{0.327}} & 0.309 & 0.345 & 0.359 & 0.341 \\
\multicolumn{1}{c|}{\multirow{-4}{*}{ETTm2}} & 720 & {\color[HTML]{FF0000} \textbf{0.386}} & {\color[HTML]{0000FF} {\ul 0.393}} & 0.426 & 0.417 & 0.436 & 0.440 & 0.433 & 0.427 & 0.417 & {\color[HTML]{FF0000} \textbf{0.390}} & {\color[HTML]{FF0000} \textbf{0.386}} & 0.395 & 0.452 & 0.419 \\ \hline
\multicolumn{1}{c|}{} & 96 & {\color[HTML]{FF0000} \textbf{0.135}} & {\color[HTML]{FF0000} \textbf{0.231}} & 0.139 & 0.237 & 0.150 & 0.253 & 0.140 & 0.238 & 0.139 & 0.241 & 0.178 & 0.276 & {\color[HTML]{0000FF} {\ul 0.138}} & {\color[HTML]{0000FF} {\ul 0.235}} \\
\multicolumn{1}{c|}{} & 192 & {\color[HTML]{0000FF} {\ul 0.152}} & {\color[HTML]{FF0000} \textbf{0.246}} & 0.156 & 0.252 & 0.164 & 0.264 & 0.160 & 0.255 & {\color[HTML]{FF0000} \textbf{0.151}} & {\color[HTML]{0000FF} {\ul 0.248}} & 0.198 & 0.293 & 0.158 & 0.255 \\
\multicolumn{1}{c|}{} & 336 & {\color[HTML]{0000FF} {\ul 0.173}} & {\color[HTML]{FF0000} \textbf{0.267}} & 0.175 & {\color[HTML]{0000FF} {\ul 0.270}} & 0.181 & 0.282 & 0.180 & 0.276 & {\color[HTML]{FF0000} \textbf{0.169}} & {\color[HTML]{0000FF} {\ul 0.270}} & 0.209 & 0.309 & 0.176 & 0.275 \\
\multicolumn{1}{c|}{\multirow{-4}{*}{ECL}} & 720 & {\color[HTML]{0000FF} {\ul 0.229}} & {\color[HTML]{0000FF} {\ul 0.312}} & 0.233 & 0.317 & {\color[HTML]{FF0000} \textbf{0.223}} & 0.321 & 0.241 & 0.323 & 0.240 & 0.322 & 0.279 & 0.355 & 0.230 & {\color[HTML]{FF0000} \textbf{0.311}} \\ \hline
\multicolumn{1}{c|}{} & 96 & {\color[HTML]{FF0000} \textbf{0.402}} & {\color[HTML]{FF0000} \textbf{0.288}} & 0.414 & 0.297 & 0.419 & 0.298 & {\color[HTML]{0000FF} {\ul 0.403}} & {\color[HTML]{0000FF} {\ul 0.289}} & 0.418 & 0.291 & 0.476 & 0.343 & 0.415 & 0.317 \\
\multicolumn{1}{c|}{} & 192 & 0.416 & {\color[HTML]{FF0000} \textbf{0.294}} & 0.426 & 0.301 & 0.434 & 0.305 & {\color[HTML]{0000FF} {\ul 0.415}} & {\color[HTML]{0000FF} {\ul 0.296}} & {\color[HTML]{FF0000} \textbf{0.414}} & {\color[HTML]{0000FF} {\ul 0.296}} & 0.496 & 0.355 & 0.425 & 0.300 \\
\multicolumn{1}{c|}{} & 336 & 0.429 & {\color[HTML]{FF0000} \textbf{0.302}} & 0.434 & {\color[HTML]{0000FF} {\ul 0.303}} & 0.449 & 0.313 & {\color[HTML]{0000FF} {\ul 0.426}} & 0.304 & {\color[HTML]{FF0000} \textbf{0.421}} & 0.311 & 0.503 & 0.356 & 0.436 & 0.310 \\
\multicolumn{1}{c|}{\multirow{-4}{*}{Traffic}} & 720 & 0.480 & {\color[HTML]{FF0000} \textbf{0.326}} & 0.487 & 0.337 & 0.484 & 0.336 & {\color[HTML]{0000FF} {\ul 0.474}} & 0.331 & {\color[HTML]{FF0000} \textbf{0.462}} & {\color[HTML]{0000FF} {\ul 0.327}} & 0.538 & 0.376 & 0.489 & 0.338 \\ \hline
\multicolumn{2}{c|}{Avg. Rank} & {\color[HTML]{FF0000} \textbf{2.036}} & {\color[HTML]{FF0000} \textbf{1.679}} & 4.000 & 3.786 & 5.143 & 5.679 & 5.036 & 5.071 & {\color[HTML]{0000FF} {\ul 2.214}} & {\color[HTML]{0000FF} {\ul 2.786}} & 4.786 & 4.821 & 4.536 & 3.857 \\ \hline
\end{tabular}
\end{table}
\renewcommand{\arraystretch}{1.1}  
\setlength{\tabcolsep}{1.5pt}       

\begin{table}[h]
\centering
\small
\caption{\redtext{\textbf{Long-term forecasting for multivariate time-series data.} We use prediction lengths \(T \in \{96,192,336,720\}\) for all datasets. The best results are in {\color[HTML]{FF0000} \textbf{bold}}, while the second-best results are {\color[HTML]{0000FF} {\ul underlined}}. Note that TEMPO* is evaluated under a zero-shot setting as it is a prompt-based method.}}
\label{tab:long_term_forecasting}
\begin{tabular}{cc|cc|cc|cc|cc|cc|cc|cc}
\hline
\multicolumn{2}{c|}{Methods} & \multicolumn{2}{c|}{LLM4TS} & \multicolumn{2}{c|}{GPT4TS} & \multicolumn{2}{c|}{DLinear} & \multicolumn{2}{c|}{PatchTST} & \multicolumn{2}{c|}{Time-LLM} & \multicolumn{2}{c|}{TEMPO*} & \multicolumn{2}{c}{TEST} \\ \hline
\multicolumn{2}{c|}{Metric} & MSE & MAE & MSE & MAE & MSE & MAE & MSE & MAE & MSE & MAE & MSE & MAE & MSE & MAE \\ \hline
\multicolumn{1}{c|}{} & 96 & {\color[HTML]{FF0000} \textbf{0.147}} & {\color[HTML]{FF0000} \textbf{0.196}} & 0.162 & 0.212 & 0.176 & 0.237 & 0.149 & {\color[HTML]{0000FF} {\ul 0.198}} & {\color[HTML]{FF0000} \textbf{0.147}} & 0.201 & 0.211 & 0.254 & 0.150 & 0.202 \\
\multicolumn{1}{c|}{} & 192 & {\color[HTML]{0000FF} {\ul 0.191}} & {\color[HTML]{0000FF} {\ul 0.238}} & 0.204 & 0.248 & 0.220 & 0.282 & 0.194 & 0.241 & {\color[HTML]{FF0000} \textbf{0.189}} & {\color[HTML]{FF0000} \textbf{0.234}} & 0.254 & 0.298 & 0.198 & 0.246 \\
\multicolumn{1}{c|}{} & 336 & {\color[HTML]{FF0000} \textbf{0.241}} & {\color[HTML]{FF0000} \textbf{0.277}} & 0.254 & 0.286 & 0.265 & 0.319 & {\color[HTML]{0000FF} {\ul 0.245}} & 0.282 & 0.262 & {\color[HTML]{0000FF} {\ul 0.279}} & 0.292 & 0.332 & {\color[HTML]{0000FF} {\ul 0.245}} & 0.286 \\
\multicolumn{1}{c|}{\multirow{-4}{*}{Weather}} & 720 & {\color[HTML]{0000FF} {\ul 0.313}} & {\color[HTML]{0000FF} {\ul 0.329}} & 0.326 & 0.337 & 0.333 & 0.362 & 0.314 & 0.334 & {\color[HTML]{FF0000} \textbf{0.304}} & {\color[HTML]{FF0000} \textbf{0.316}} & 0.370 & 0.379 & 0.324 & 0.342 \\ \hline
\multicolumn{1}{c|}{} & 96 & 0.371 & {\color[HTML]{0000FF} {\ul 0.394}} & 0.376 & 0.397 & 0.375 & 0.399 & {\color[HTML]{0000FF} {\ul 0.370}} & 0.399 & {\color[HTML]{FF0000} \textbf{0.362}} & {\color[HTML]{FF0000} \textbf{0.392}} & 0.400 & 0.406 & 0.372 & 0.400 \\
\multicolumn{1}{c|}{} & 192 & {\color[HTML]{0000FF} {\ul 0.403}} & {\color[HTML]{FF0000} \textbf{0.412}} & 0.416 & 0.418 & 0.405 & {\color[HTML]{0000FF} {\ul 0.416}} & 0.413 & 0.421 & {\color[HTML]{FF0000} \textbf{0.398}} & 0.418 & 0.426 & 0.421 & 0.414 & 0.422 \\
\multicolumn{1}{c|}{} & 336 & {\color[HTML]{FF0000} \textbf{0.420}} & {\color[HTML]{FF0000} \textbf{0.422}} & 0.442 & 0.433 & 0.439 & 0.443 & {\color[HTML]{0000FF} {\ul 0.422}} & 0.436 & 0.430 & {\color[HTML]{0000FF} {\ul 0.427}} & 0.441 & 0.430 & {\color[HTML]{0000FF} {\ul 0.422}} & 0.437 \\
\multicolumn{1}{c|}{\multirow{-4}{*}{ETTh1}} & 720 & {\color[HTML]{FF0000} \textbf{0.422}} & {\color[HTML]{FF0000} \textbf{0.444}} & 0.477 & 0.456 & 0.472 & 0.490 & 0.447 & 0.466 & {\color[HTML]{0000FF} {\ul 0.442}} & 0.457 & 0.443 & {\color[HTML]{0000FF} {\ul 0.451}} & 0.447 & 0.467 \\ \hline
\multicolumn{1}{c|}{} & 96 & {\color[HTML]{0000FF} {\ul 0.269}} & {\color[HTML]{0000FF} {\ul 0.332}} & 0.285 & 0.342 & 0.289 & 0.353 & 0.274 & 0.336 & {\color[HTML]{FF0000} \textbf{0.268}} & {\color[HTML]{FF0000} \textbf{0.328}} & 0.301 & 0.353 & 0.275 & 0.338 \\
\multicolumn{1}{c|}{} & 192 & {\color[HTML]{FF0000} \textbf{0.328}} & {\color[HTML]{0000FF} {\ul 0.377}} & 0.354 & 0.389 & 0.383 & 0.418 & 0.339 & 0.379 & {\color[HTML]{0000FF} {\ul 0.329}} & {\color[HTML]{FF0000} \textbf{0.375}} & 0.355 & 0.389 & 0.340 & 0.379 \\
\multicolumn{1}{c|}{} & 336 & 0.353 & 0.396 & 0.373 & 0.407 & 0.448 & 0.465 & {\color[HTML]{FF0000} \textbf{0.329}} & {\color[HTML]{FF0000} \textbf{0.380}} & 0.368 & 0.409 & 0.379 & 0.408 & {\color[HTML]{FF0000} \textbf{0.329}} & {\color[HTML]{0000FF} {\ul 0.381}} \\
\multicolumn{1}{c|}{\multirow{-4}{*}{ETTh2}} & 720 & 0.383 & 0.425 & 0.406 & 0.441 & 0.605 & 0.551 & {\color[HTML]{0000FF} {\ul 0.379}} & {\color[HTML]{0000FF} {\ul 0.422}} & {\color[HTML]{FF0000} \textbf{0.372}} & {\color[HTML]{FF0000} \textbf{0.420}} & 0.409 & 0.440 & 0.381 & 0.423 \\ \hline
\multicolumn{1}{c|}{} & 96 & {\color[HTML]{0000FF} {\ul 0.285}} & 0.343 & 0.292 & 0.346 & 0.299 & 0.343 & 0.290 & {\color[HTML]{0000FF} {\ul 0.342}} & {\color[HTML]{FF0000} \textbf{0.272}} & {\color[HTML]{FF0000} \textbf{0.334}} & 0.438 & 0.424 & 0.293 & 0.346 \\
\multicolumn{1}{c|}{} & 192 & {\color[HTML]{0000FF} {\ul 0.324}} & 0.366 & 0.332 & 0.372 & 0.335 & {\color[HTML]{0000FF} {\ul 0.365}} & 0.332 & 0.369 & {\color[HTML]{FF0000} \textbf{0.310}} & {\color[HTML]{FF0000} \textbf{0.358}} & 0.461 & 0.432 & 0.332 & 0.369 \\
\multicolumn{1}{c|}{} & 336 & {\color[HTML]{0000FF} {\ul 0.353}} & {\color[HTML]{0000FF} {\ul 0.385}} & 0.366 & 0.394 & 0.369 & 0.386 & 0.366 & 0.392 & {\color[HTML]{FF0000} \textbf{0.352}} & {\color[HTML]{FF0000} \textbf{0.384}} & 0.515 & 0.467 & 0.368 & 0.392 \\
\multicolumn{1}{c|}{\multirow{-4}{*}{ETTm1}} & 720 & {\color[HTML]{0000FF} {\ul 0.408}} & {\color[HTML]{0000FF} {\ul 0.419}} & 0.417 & 0.421 & 0.425 & 0.421 & 0.416 & 0.420 & {\color[HTML]{FF0000} \textbf{0.383}} & {\color[HTML]{FF0000} \textbf{0.411}} & 0.591 & 0.509 & 0.418 & 0.420 \\ \hline
\multicolumn{1}{c|}{} & 96 & {\color[HTML]{0000FF} {\ul 0.165}} & 0.254 & 0.173 & 0.262 & 0.167 & 0.269 & {\color[HTML]{0000FF} {\ul 0.165}} & 0.255 & {\color[HTML]{FF0000} \textbf{0.161}} & {\color[HTML]{0000FF} {\ul 0.253}} & 0.185 & 0.267 & 0.193 & {\color[HTML]{FF0000} \textbf{0.237}} \\
\multicolumn{1}{c|}{} & 192 & {\color[HTML]{0000FF} {\ul 0.220}} & {\color[HTML]{0000FF} {\ul 0.292}} & 0.229 & 0.301 & 0.224 & 0.303 & {\color[HTML]{0000FF} {\ul 0.220}} & {\color[HTML]{0000FF} {\ul 0.292}} & {\color[HTML]{FF0000} \textbf{0.219}} & 0.293 & 0.243 & 0.304 & 0.257 & {\color[HTML]{FF0000} \textbf{0.264}} \\
\multicolumn{1}{c|}{} & 336 & {\color[HTML]{FF0000} \textbf{0.268}} & {\color[HTML]{0000FF} {\ul 0.326}} & 0.286 & 0.341 & 0.281 & 0.342 & 0.274 & 0.329 & {\color[HTML]{0000FF} {\ul 0.271}} & 0.329 & 0.309 & 0.345 & 0.289 & {\color[HTML]{FF0000} \textbf{0.295}} \\
\multicolumn{1}{c|}{\multirow{-4}{*}{ETTm2}} & 720 & {\color[HTML]{FF0000} \textbf{0.350}} & 0.380 & 0.378 & 0.401 & 0.397 & 0.421 & 0.362 & 0.385 & {\color[HTML]{0000FF} {\ul 0.352}} & {\color[HTML]{0000FF} {\ul 0.379}} & 0.386 & 0.395 & 0.375 & {\color[HTML]{FF0000} \textbf{0.369}} \\ \hline
\multicolumn{1}{c|}{} & 96 & {\color[HTML]{FF0000} \textbf{0.128}} & {\color[HTML]{0000FF} {\ul 0.223}} & 0.139 & 0.238 & 0.140 & 0.237 & {\color[HTML]{0000FF} {\ul 0.129}} & {\color[HTML]{FF0000} \textbf{0.222}} & 0.131 & 0.224 & 0.178 & 0.276 & 0.132 & {\color[HTML]{0000FF} {\ul 0.223}} \\
\multicolumn{1}{c|}{} & 192 & {\color[HTML]{FF0000} \textbf{0.146}} & {\color[HTML]{FF0000} \textbf{0.240}} & 0.153 & 0.251 & 0.153 & 0.249 & 0.157 & {\color[HTML]{FF0000} \textbf{0.240}} & {\color[HTML]{0000FF} {\ul 0.152}} & 0.241 & 0.198 & 0.293 & 0.158 & 0.241 \\
\multicolumn{1}{c|}{} & 336 & {\color[HTML]{0000FF} {\ul 0.163}} & {\color[HTML]{0000FF} {\ul 0.258}} & 0.169 & 0.266 & 0.169 & 0.267 & {\color[HTML]{0000FF} {\ul 0.163}} & 0.259 & {\color[HTML]{FF0000} \textbf{0.160}} & {\color[HTML]{FF0000} \textbf{0.248}} & 0.209 & 0.309 & {\color[HTML]{0000FF} {\ul 0.163}} & 0.260 \\
\multicolumn{1}{c|}{\multirow{-4}{*}{ECL}} & 720 & 0.200 & 0.292 & 0.206 & 0.297 & 0.203 & 0.301 & {\color[HTML]{0000FF} {\ul 0.197}} & {\color[HTML]{FF0000} \textbf{0.290}} & {\color[HTML]{FF0000} \textbf{0.192}} & 0.298 & 0.279 & 0.355 & 0.199 & {\color[HTML]{0000FF} {\ul 0.291}} \\ \hline
\multicolumn{1}{c|}{} & 96 & 0.372 & 0.259 & 0.388 & 0.282 & 0.410 & 0.282 & {\color[HTML]{FF0000} \textbf{0.360}} & {\color[HTML]{0000FF} {\ul 0.249}} & {\color[HTML]{0000FF} {\ul 0.362}} & {\color[HTML]{FF0000} \textbf{0.248}} & 0.476 & 0.343 & 0.407 & 0.282 \\
\multicolumn{1}{c|}{} & 192 & 0.391 & 0.265 & 0.407 & 0.290 & 0.423 & 0.287 & {\color[HTML]{0000FF} {\ul 0.379}} & {\color[HTML]{0000FF} {\ul 0.256}} & {\color[HTML]{FF0000} \textbf{0.374}} & {\color[HTML]{FF0000} \textbf{0.247}} & 0.496 & 0.355 & 0.423 & 0.287 \\
\multicolumn{1}{c|}{} & 336 & 0.405 & 0.275 & 0.412 & 0.294 & 0.436 & 0.296 & {\color[HTML]{0000FF} {\ul 0.392}} & {\color[HTML]{FF0000} \textbf{0.264}} & {\color[HTML]{FF0000} \textbf{0.385}} & {\color[HTML]{0000FF} {\ul 0.271}} & 0.503 & 0.356 & 0.430 & 0.296 \\
\multicolumn{1}{c|}{\multirow{-4}{*}{Traffic}} & 720 & 0.437 & 0.292 & 0.450 & 0.312 & 0.466 & 0.315 & {\color[HTML]{0000FF} {\ul 0.432}} & {\color[HTML]{FF0000} \textbf{0.286}} & {\color[HTML]{FF0000} \textbf{0.430}} & {\color[HTML]{0000FF} {\ul 0.288}} & 0.538 & 0.376 & 0.463 & 0.315 \\ \hline
\multicolumn{2}{c|}{Avg. Rank} & {\color[HTML]{0000FF} {\ul 2.036}} & {\color[HTML]{0000FF} {\ul 2.214}} & 4.786 & 4.750 & 5.536 & 5.357 & 2.571 & 2.750 & {\color[HTML]{FF0000} \textbf{1.643}} & {\color[HTML]{FF0000} \textbf{2.143}} & 6.607 & 6.250 & 4.214 & 3.679 \\ \hline
\end{tabular}
\end{table}

\paragraph{\textbf{Implementation Details}}
For our experiments in long-term time-series forecasting, few-shot learning, and ablation studies, we utilize the settings from PatchTST \cite{patchtst} for a consistent comparison.
We first explore the model's performance in few-shot scenarios, followed by a comprehensive evaluation under full-shot settings to ensure a thorough and balanced analysis.
We set our look-back window length \(T_{in}\) to either \(336\) or \(512\) (reporting the best results), and configure the patch length \(P\) as \(16\) with a stride \(S\) of \(8\).
For unsupervised representation learning, the settings are slightly adjusted to \(T_{in} = 512\), \(P=12\), and \(S = 12\).
Aligned with the GPT4TS configuration \cite{gpt4ts}, we utilize only the first 6 layers of the 12-layer GPT-2 base \cite{gpt2}.

\subsection{Few-Shot Learning in Long-Term Time-Series Forecasting}
Table \ref{tab:few_shot_learning_5} shows the results of long-term time-series forecasting using only \(5\%\) of the training data, while Table \ref{tab:few_shot_learning_10} presents similar results but with merely \(10\%\) of the training data utilized.
In our experiments, consistent splits for training, validation, and test sets are maintained across both full and few-shot learning scenarios.
We deliberately limited the training data percentage to 5\% and 10\% to evaluate model performance in few-shot scenarios.
For each dataset, we train a single model in the time-series alignment stage, which is then applied consistently across all prediction lengths.
In contrast, in the forecasting fine-tuning stage, we fine-tune a distinct model for each prediction length, while ensuring that all these models share the same hyperparameters.

Both LLM4TS and GPT4TS \cite{gpt4ts} consistently surpass most train-from-scratch models in limited-data scenarios across various datasets, thanks to the pre-existing representation learning capability encapsulated in GPT-2.
With the additional time-series alignment and multi-scale temporal information integration, LLM4TS emerges as a better data-efficient time-series forecaster against GPT4TS, achieving better performance across all datasets.
Notably, LLM4TS with only \(5\%\) of data outperforms the best baseline that uses \(10\%\) of data.
For the largest dataset (Traffic), PatchTST emerges as the leading model in the full-shot scenario, though this trend does not extend to few-shot scenarios.
With only \(10\%\) training data, LLM4TS outperforms PatchTST in 5 out of 8 evaluations, and with just \(5\%\) training data, it leads in 5 out of 6 evaluations.
This suggests that in few-shot scenarios, traditional deep train-from-scratch models generally still underperform compared to those leveraging pre-trained LLMs.

\redtext{
We also compare the latest LLM-based methods in the domain of time-series forecasting: Time-LLM, TEMPO, and TEST.
It’s important to note that TEMPO is evaluated under a zero-shot setting, as it is a prompt-based method.
Consequently, in both few-shot and full-shot scenarios, TEMPO consistently reports zero-shot results.
For larger datasets (like Weather, Electricity, and Traffic), LLM4TS consistently achieves the best performance, even with only 5\% or 10\% of the training data.
TEMPO performs best on the ETTh1 and ETTh2 datasets, while Time-LLM leads on the ETTm1 and ETTm2 datasets.
Overall, LLM4TS demonstrates superior performance across all these LLM-based methods in both the 5\% and 10\% training data scenarios.
}

\subsection{Full-Shot Learning in Long-Term Time-Series Forecasting}
Table \ref{tab:long_term_forecasting} presents the results of long-term time-series forecasting averaged over a consistent prediction length set \(T_{out} \in \{96,192,336,720\}\) for all datasets.
While the primary focus of using pre-trained LLMs is on few-shot learning, LLM4TS not only excels in this area but also outperforms all deep train-from-scratch methods, even with full dataset access, thanks to its two-stage fine-tuning and integration of multi-scale temporal information.
In contrast, GPT4TS, despite utilizing the pre-trained GPT-2's representation learning capabilities, does not achieve superior performance over traditional train-from-scratch baselines in full-shot scenarios.
This is particularly evident when handling extensive volumes of training data.
This limitation mainly stems from its absence of time-series alignment and the inclusion of multi-scale temporal information, which is essential for enhancing performance in time-series forecasting.
Interestingly, for the largest dataset (Traffic), PatchTST can outcompete both LLM4TS and GPT4TS.
This suggests that with complete dataset access and sufficient data volume, traditional deep train-from-scratch models may sometimes outshine those leveraging pre-trained LLMs.

\redtext{
We also compare the latest LLM-based methods in the domain of time-series forecasting: Time-LLM, TEMPO, and TEST.
It’s important to note that TEMPO is evaluated under a zero-shot setting, as it is a prompt-based method, and therefore consistently reports zero-shot results in both few-shot and full-shot scenarios.
Time-LLM emerges as the leading model, thanks to its innovative approach of reprogramming LLMs for time series forecasting by converting time series into text prototypes and using prompts to guide predictions.
However, LLM4TS remains very close in performance to Time-LLM.
Notably, LLM4TS still outperforms Time-LLM in few-shot scenarios, demonstrating its exceptional data efficiency and robustness in limited-data settings.
}

\renewcommand{\arraystretch}{1.4}  
\setlength{\tabcolsep}{4pt}       

\begin{table}[t]
\small
\centering
\caption{\textbf{Unsupervised representation learning evaluation in forecasting with linear probing.} We use prediction lengths \(T_{out} \in \{24, 48, 168, 336, 720\}\) for the ETTh1 dataset. The best average results are in {\color[HTML]{FF0000} \textbf{bold}}, while the second-best results are in {\color[HTML]{0000FF} {\ul underlined}}.}
\label{tab:self_supervised}
\begin{tabular}{cc|cc|cc|cc|cc|cc|cc}
\hline
\multicolumn{2}{c|}{Methods} & \multicolumn{2}{c|}{LLM4TS} & \multicolumn{2}{c|}{PatchTST} & \multicolumn{2}{c|}{BTSF} & \multicolumn{2}{c|}{TS2Vec} & \multicolumn{2}{c|}{TNC} & \multicolumn{2}{c}{TS-TCC} \\ \hline
\multicolumn{2}{c|}{Metric} & MSE & MAE & MSE & MAE & MSE & MAE & MSE & MAE & MSE & MAE & MSE & MAE \\ \hline
\multicolumn{1}{c|}{} & 24 & 0.315 & 0.365 & 0.322 & 0.369 & 0.541 & 0.519 & 0.599 & 0.534 & 0.632 & 0.596 & 0.653 & 0.610 \\
\multicolumn{1}{c|}{} & 48 & 0.342 & 0.384 & 0.354 & 0.385 & 0.613 & 0.524 & 0.629 & 0.555 & 0.705 & 0.688 & 0.720 & 0.693 \\
\multicolumn{1}{c|}{} & 168 & 0.401 & 0.415 & 0.419 & 0.424 & 0.640 & 0.532 & 0.755 & 0.636 & 1.097 & 0.993 & 1.129 & 1.044 \\
\multicolumn{1}{c|}{} & 336 & 0.421 & 0.427 & 0.445 & 0.446 & 0.864 & 0.689 & 0.907 & 0.717 & 1.454 & 0.919 & 1.492 & 1.076 \\
\multicolumn{1}{c|}{} & 720 & 0.426 & 0.447 & 0.487 & 0.478 & 0.993 & 0.712 & 1.048 & 0.790 & 1.604 & 1.118 & 1.603 & 1.206 \\
\multicolumn{1}{c|}{\multirow{-6}{*}{ETTh1}} & Avg. & {\color[HTML]{FF0000} \textbf{0.381}} & {\color[HTML]{FF0000} \textbf{0.408}} & {\color[HTML]{0000FF} {\ul 0.405}} & {\color[HTML]{0000FF} {\ul 0.420}} & 0.730 & 0.595 & 0.788 & 0.646 & 1.098 & 0.863 & 1.119 & 0.926 \\ \hline
\end{tabular}
\end{table}

\subsection{Unsupervised Representation Learning}
Given that the autoregressive objective used in the time-series alignment stage can be seen as a pretext task in unsupervised representation learning, we aim to assess LLM4TS's representation learning capability.
To evaluate the effectiveness of unsupervised representation learning, we conduct a linear evaluation on time-series forecasting.
This involves pre-training the backbone model using the pretext task, freezing its weights, and then training an attached linear layer on the downstream forecasting task.
With the backbone model's parameters fixed, strong performance in forecasting depends on the expressiveness of the learned representations.
Table \ref{tab:self_supervised} shows LLM4TS's superior performance over competitors on the ETTh1 dataset, highlighting the effectiveness of adapting the LLM to time-series characteristics in the time-series alignment stage.
This comparison exclusively includes self-supervised learning methods, thereby excluding deep train-from-scratch models designed explicitly for time-series forecasting.
Similarly, GPT4TS is not part of this experiment as it lacks a distinct stage of representation learning.
The variant of PatchTST used here differs from that in the forecasting experiments; this variant focuses on representation learning.
PatchTST incorporates an MLM (Masked Language Model) approach, akin to BERT, for learning representations.
Despite this, LLM4TS still emerges as the top method in representation learning capability among all evaluated methods, achieving an average improvement of \(6.02\%\) in MSE.
Within the domain of unsupervised representation learning for time-series data, models based on CNN (such as BTSF, TS2Vec, TS-TCC) and RNN (like TNC) are typically favored over Transformers.
However, our experiments reveal that Transformer-based models can surpass these traditional choices in performance with well-designed pretext tasks.

\begin{figure}[t]
    \centering
    \includegraphics[width=1\linewidth]{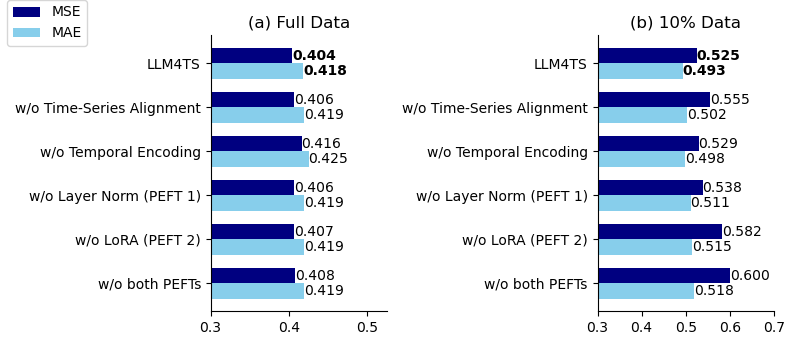}
    \caption{\redtext{\textbf{Ablation study on key components in LLM4TS.} Each ablation is conducted under both full- and few-shot learning with \(10\%\) training data. We report results averaged over prediction lengths \(T_{out} \in \{96, 192, 336, 720\}\) for the ETTh1 dataset. The best results are in \textbf{bold}.}}
    \label{fig:ablation_part_a}
\end{figure}

\subsection{Ablation Study}
\subsubsection{Key Components in LLM4TS}
Fig. \ref{fig:ablation_part_a} explores the effects of time-series alignment, multi-scale temporal encoding, and PEFT in LLM4TS, assessing both full- and few-shot scenarios on the ETTh1 dataset.
A comparative analysis—with and without these components—highlights their individual importance in enhancing forecasting accuracy in both scenarios.
Notably, LLM4TS delivers exceptional performance in few-shot learning, averaging a \(6.2\%\) reduction in MSE with each incorporation of these components.

In the experimental results, we observe three key insights.
First, there is a notable trend where the MSE improvement increases as the prediction length extends.
This indicates that the core elements of LLM4TS become increasingly beneficial in situations where a higher level of predictive capability is required, particularly evident with longer prediction lengths.
Second, few-shot scenarios exhibit more substantial gains than full-shot scenarios upon integrating these main components into LLM4TS.
It emphasizes LLM4TS's strength as a data-efficient time-series forecaster, a quality primarily attributed to its intrinsic components.
\redtext{
Third, among the two PEFT methods, LoRA proves to be more beneficial than Layer Normalization.
This advantage is consistently observed in both full-shot and few-shot scenarios, highlighting LoRA's effectiveness in enhancing the model's performance. 
}

\subsubsection{Training Strategies in Forecasting Fine-Tuning}
\begin{figure}[t]
    \centering
    \includegraphics[width=1\linewidth]{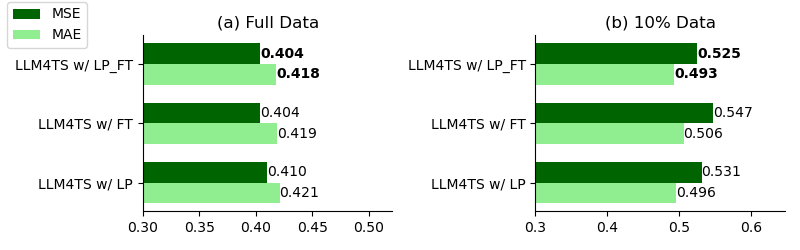}
    \caption{\textbf{Ablation study on training strategies in forecasting fine-tuning.} Each ablation is conducted under both full- and few-shot learning with \(10\%\) training data. We report results averaged over prediction lengths \(T_{out} \in \{96, 192, 336, 720\}\) for the ETTh1 dataset. The best results are in \textbf{bold}.}
    \label{fig:ablation_part_b}
\end{figure}

\redtext{As discussed in Section \ref{sec:forecasting_fine_tuning}, while linear probing (LP) shows superior performance in out-of-distribution (OOD) scenarios and full fine-tuning (FT) excels in in-distribution (ID) scenarios, LP-FT can surpass FT and LP in both OOD and ID scenarios.}
Fig. \ref{fig:ablation_part_b} shows that LP-FT enhances performance in both full- and few-shot learning on the ETTh1 dataset, achieving an average improvement of \(0.7\%\) in MSE for full-shot learning and \(2.51\%\) for few-shot learning.
The subtle improvements in both scenarios can be attributed to the limited number of trainable parameters in the LLM4TS's backbone model even when using FT, which narrows the distinction between LP and FT.
The results further reveal that few-shot learning derives a greater advantage from LP-FT, primarily due to its higher vulnerability to severe OOD issues.
Additionally, consistent with observations in the ablation study of LLM4TS's main components, we note a similar trend where longer prediction lengths yield more significant benefits in few-shot scenarios.

\renewcommand{\arraystretch}{1.2}  
\setlength{\tabcolsep}{10pt}       

\begin{table}[h]
\centering
\small
\caption{\textbf{Ablation study on the effectiveness of LLM's pre-trained weights}. Each ablation is conducted under few-shot learning with 10\% and 5\% training data. 'No Freeze' refers to the model utilizing LLM's pre-trained weights without freezing any layers during training, whereas 'No Pretrain' denotes the model not utilizing LLM's pre-trained weights, implying the model is trained from scratch. We report results averaged over prediction lengths \(T_{out} \in \{96,192,336,720\}\) for the Weather, ETTm1, and ETTm2 datasets. The best average results are in {\color[HTML]{FF0000} \textbf{bold}}.}
\label{tab:ablation_part_c}
\begin{tabular}{cc|cc|cc|cc|cc}
\hline
\multicolumn{2}{c|}{Methods} & \multicolumn{2}{c|}{LLM4TS} & \multicolumn{2}{c|}{GPT4TS} & \multicolumn{2}{c|}{No Freeze} & \multicolumn{2}{c}{No Pretrain} \\ \hline
\multicolumn{2}{c|}{Metric} & MSE & MAE & MSE & MAE & MSE & MAE & MSE & MAE \\ \hline
\multicolumn{1}{c|}{} & Weather & {\color[HTML]{FF0000} \textbf{0.235}} & {\color[HTML]{FF0000} \textbf{0.270}} & 0.238 & 0.275 & 0.273 & 0.302 & 0.278 & 0.305 \\
\multicolumn{1}{c|}{} & ETTm1 & {\color[HTML]{FF0000} \textbf{0.408}} & {\color[HTML]{FF0000} \textbf{0.413}} & 0.464 & 0.441 & 0.546 & 0.484 & 0.473 & 0.446 \\
\multicolumn{1}{c|}{\multirow{-3}{*}{10\%}} & ETTm2 & {\color[HTML]{FF0000} \textbf{0.276}} & {\color[HTML]{FF0000} \textbf{0.324}} & 0.293 & 0.335 & 0.340 & 0.367 & 0.361 & 0.385 \\ \hline
\multicolumn{1}{c|}{} & Weather & {\color[HTML]{FF0000} \textbf{0.256}} & {\color[HTML]{FF0000} \textbf{0.292}} & 0.264 & 0.302 & 0.284 & 0.312 & 0.298 & 0.324 \\
\multicolumn{1}{c|}{} & ETTm1 & {\color[HTML]{FF0000} \textbf{0.413}} & {\color[HTML]{FF0000} \textbf{0.417}} & 0.467 & 0.450 & 0.562 & 0.496 & 0.470 & 0.452 \\
\multicolumn{1}{c|}{\multirow{-3}{*}{5\%}} & ETTm2 & {\color[HTML]{FF0000} \textbf{0.286}} & {\color[HTML]{FF0000} \textbf{0.332}} & 0.308 & 0.347 & 0.327 & 0.362 & 0.413 & 0.411 \\ \hline
\end{tabular}
\end{table}

\subsubsection{Effectiveness of LLM's Pre-trained Weights}
\label{sec:ablation_part_c}
As discussed in Section \ref{sec:transfer_learning_with_llms}, most parameters in these LLMs are kept fixed to preserve their data-independent representation learning capability.
Table \ref{tab:ablation_part_c} shows that LLM4TS performs the best when most parameters remain unchanged on the Weather, ETTm1, and ETTm2 datasets.
Specifically, LLM4TS demonstrates a notable average improvement of \(17.78\%\) in MSE compared to the 'No Freeze' approach, where pre-trained weights are utilized without freezing any layers during training.
Furthermore, when compared to the 'No Pretrain' approach, where the model is trained from scratch without leveraging pre-trained weights, LLM4TS showcases an even more significant average improvement of \(18.28\%\) in MSE. 
This emphasizes the importance of retaining the pre-existing strengths in representation learning inherent to these models, attributable primarily to the self-attention mechanisms within pre-trained transformers, which facilitate the development of data-independent operations.

\renewcommand{\arraystretch}{1.2}  
\setlength{\tabcolsep}{12pt}       

\begin{table}[h]
\centering
\small
\caption{\textbf{Training parameters}.}
\label{tab:training_params}
\begin{tabular}{c|ccc}
\hline
Model & Trainable Parameters & Total Parameters & Trainable Parameters Percentage \\ \hline
LLM4TS & 3.4M & 85M & 4\% \\
PatchTST & 20M & 20M & 100\% \\
FEDformer & 33M & 33M & 100\% \\ \hline
\end{tabular}
\end{table}

\begin{figure}[t]
    \centering
    \includegraphics[width=1\linewidth]{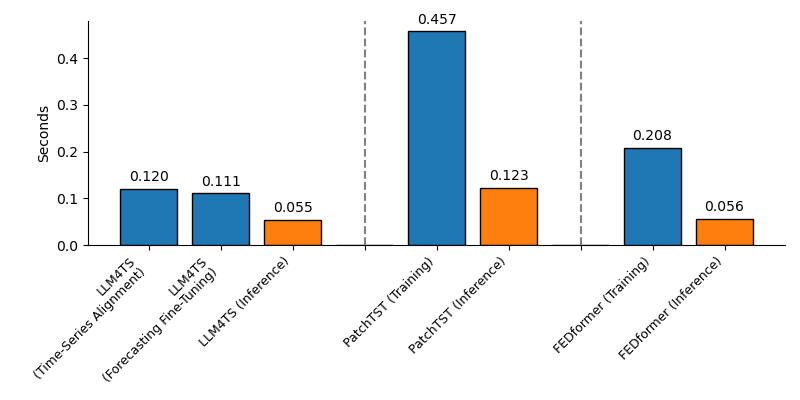}
    \caption{\redtext{\textbf{Comparison of training and inference time (in seconds) for one batch.} We use the prediction length \(T_{out} = 96\) for the ETTh2 dataset. The best results are in \textbf{bold}.}}
    \label{fig:running_time}
\end{figure}

\subsection{Training and Inference Cost}
Evaluating the computational costs of LLM-based models is essential for determining their practicality in real-world scenarios.
In this context, we compare LLM4TS with two other leading Transformer-based baselines, PatchTST and FEDformer.
Details regarding the number of trainable parameters and the total parameters can be found in Table \ref{tab:training_params}.
LLM4TS distinguishes itself by keeping most pre-trained parameters fixed and employing two Parameter-Efficient Fine-Tuning (PEFT) methods: Layer Normalization Tuning and LoRA.
Consequently, only 4\% of its parameters are trainable, rendering the number of trainable parameters in LLM4TS significantly lower than those in its train-from-scratch counterparts.

The execution time for both training and inference of LLM4TS, in comparison with PatchTST and FEDformer, is evaluated using an NVIDIA Tesla V100 GPU, and the results are illustrated in Fig. \ref{fig:running_time}.
To ensure a fair comparison across all methods, we standardized the batch size at 128 and set the hidden dimensions to 768, aligning with the specifications of GPT-2.
\redtext{
The evaluation was conducted on a single batch, and for LLM4TS, we provided the training time for both stages, as it involved a two-stage training process.
}
LLM4TS outperformed the baselines in execution time during both the training and inference stages.
This enhanced efficiency is credited to its architecture, which leverages a majority of non-trainable parameters, thus markedly diminishing the computational burden throughout both the training and inference phases.

\section{Conclusion}
In this paper, we present LLM4TS, a framework for time-series forecasting utilizing pre-trained LLMs.
LLM4TS employs a two-stage fine-tuning strategy, beginning with the \textit{time-series alignment} stage to adapt LLMs to the characteristics of time-series data, followed by the \textit{forecasting fine-tuning} stage designed for time-series forecasting tasks.
Our framework also introduces a novel two-level aggregation method, integrating multi-scale temporal data within pre-trained LLMs to improve their interpretation of time-related information.
Through experiments on 7 time-series forecasting datasets, LLM4TS demonstrates superior performance over existing state-of-the-art methods, including those trained from scratch, in both full and few-shot scenarios.

In future work, we plan to extend our research in two directions.
\redtext{
First, while we chose GPT-2 as our primary LLM in this paper for a fair comparison over GPT4TS, we plan to evaluate more recent LLMs like GPT-3.5 and LLaMA-2 to assess their advancements.  
}
Second, we aim to explore other tasks, such as classification and anomaly detection.
Although forecasting is highly relevant to real-world applications without the need for manual labeling, extending it to other tasks enables the broader applicability of our LLM4TS framework.

\bibliographystyle{plain}
\bibliography{Reference}

\end{document}